# Occupation-aware planning method for robotic monitoring missions in dynamic environments

Yaroslav Marchukov and Luis Montano

*Instituto de Investigación en Ingeniería de Aragón (I3A), University of Zaragoza, Spain*
*yamar@unizar.es, montano@unizar.es*

**Abstract**

This paper presents a method for robotic monitoring missions in the presence of moving obstacles. Although the scenario map is known, the robot lacks information about the movement of dynamic obstacles during the monitoring mission. Numerous local planners have been developed in recent years for navigating highly dynamic environments. However, the absence of a global planner for these environments can result in unavoidable collisions or the inability to successfully complete missions in densely populated areas, such as a scenario monitoring in our case. This work addresses the development and evaluation of a global planner, *MADA* (Monitoring Avoiding Dynamic Areas), aimed at enhancing the deployment of robots in such challenging conditions. The robot plans and executes the mission using the proposed two-step approach. The first step involves selecting the observation goal based on the environment's distribution and estimated monitoring costs. In the second step, the robot identifies areas with moving obstacles and obtains paths avoiding densely occupied dynamic regions based on their occupation. Quantitative and qualitative results based on simulations and on real-world experimentation, confirm that the proposed method allows the robot to effectively monitor most of the environment while avoiding densely occupied dynamic areas.

## 1. Introduction

The paper focuses in the development of a global planner for dynamic environments. Traditional planning techniques designed for static scenarios are no longer applicable, as the planned trajectories need constant re-planning to navigate through areas temporarily occupied by moving obstacles. The global planner must be developed with context awareness, tailored to the specific mission at hand. Traditional criteria such as the shortest path or minimum time to the goal are no longer universally applicable. In this work we center in environment monitoring missions, which impose some constraints and characteristics to the designed planner. The robot's mission is to observe an entire environment, given the map of the scenario containing the static obstacles (walls, furniture, etc). However, the presence of dynamic obstacles, whose motions are unknown and introduce uncertainty into the scenario, presents an additional challenge, addressed in this work. The robot must avoid colliding with them or traversing these areas, whilst keeping the main mission of monitoring the most part possible of the environment. Figure 1 illustrates two snapshots of a real-world experiment in which the proposed method has been applied. Trajectory planning for static or sporadically changing scenarios, or a purely reactive approach to obstacle avoidance, are inadequate. Even if these methods consider surrounding moving obstacles, they fall short because a new trajectory must be continuously computed to monitor the majority of the scenario at any given moment while simultaneously managing its dynamic nature. The robot tackles the challenge of balancing collision risk with mission completion. Once the scenario is observed, either completely or partially due to occlusions provoked in dynamic dense areas, the mission is considered complete. Thus, the problem solved in this work is the development of a global planner to monitor a scenario, avoiding obstacle-crowded regions.

A potential application involves monitoring factories, warehouses, or supermarkets, where a robot can be utilized to search for objects or for environment surveillance, maintaining the maximum visibility as possible

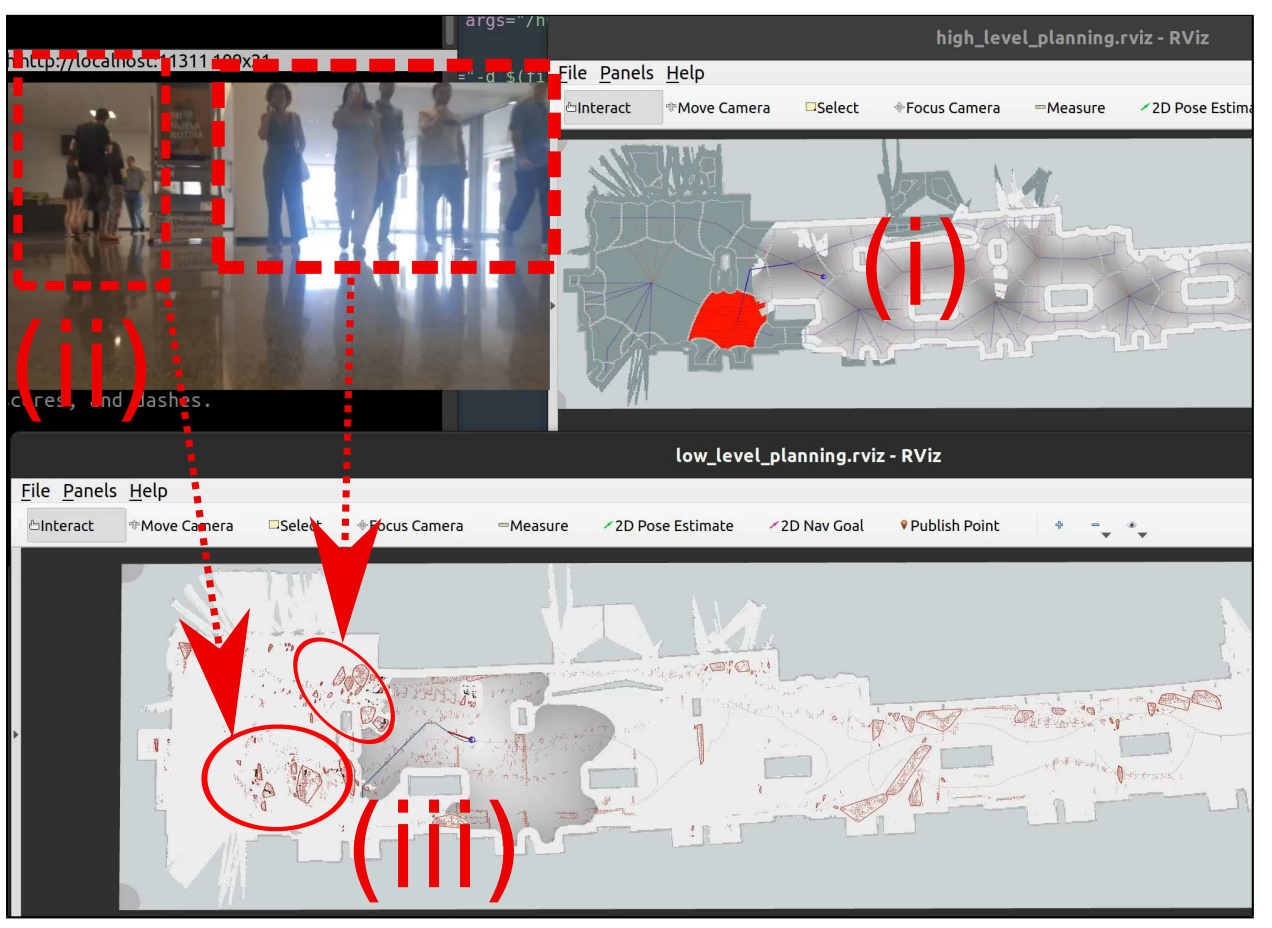

(a) Initial plan. (i) The robot computes the goal from where to monitor the next set of points (red points). (ii) There are dynamic obstacles in the scenario observed by the robot. (iii) The robot builds the dynamic areas based on their movement (red polygons) and computes an initial global plan to reach the goal (blue line).

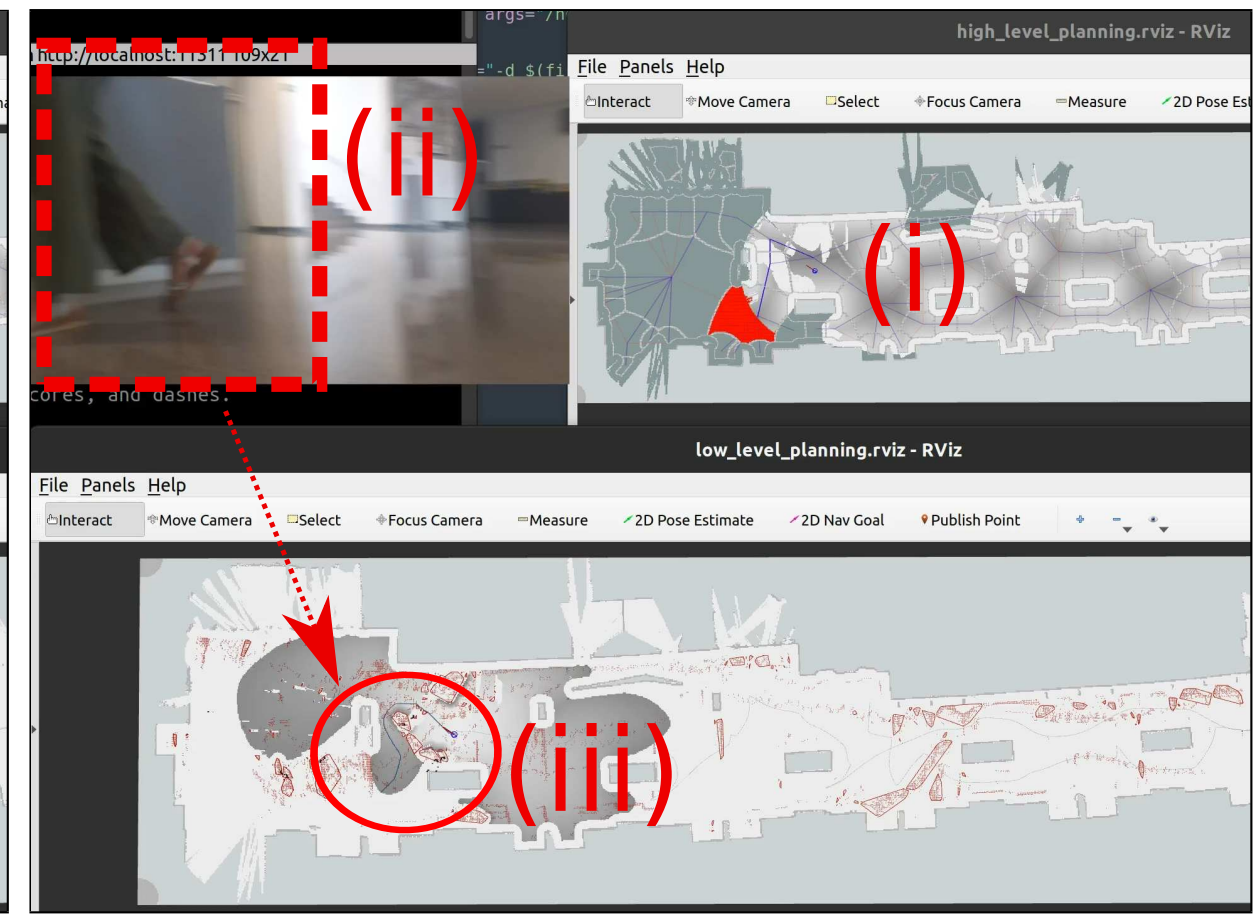

(b) Plan update based on the movement of obstacles. (i) The robot updates the monitored part of the scenario. (ii) It enlarges the dynamic areas as the obstacles move. (iii) A new plan is obtained surrounding the updated dynamic areas, so as not to disturb humans and avoiding to collide with them.

Figure 1: Monitoring mission in presence of humans.

in every moment. The presence of human workers or other robots acting as dynamic obstacles, introduces uncertainty in the global plan, necessitating its recomputation at each control interval. Additionally, in these scenarios, employees typically move within designated working areas. The robot must avoid these densely populated dynamic areas to reduce the risk of collision and ensure thorough environmental observability. There are three main reasons for the robot to avoid these areas. The first reason is to prevent any disturbance to moving obstacles. The second is to avoid collisions in populated areas, mainly in narrow and highly populated areas, in which the robot could cause significant damage to people or even break itself. The third is trying to keep the maximum visibility of the scenario, because the moving obstacles could keep unobserved a part of the monitored area. Therefore, it is crucial for the robot to identify these dynamic areas, enforcing to the motion planner to move the robot towards clearer areas that can be correctly monitored without a great reduction of the observed areas, allowing to re-visit them when possible. A conclusion of this work is that it is needed to integrate specif and context-aware global planners with efficient local planners to improve the success of the mission.

Though the monitoring application has been addressed to frame the desired robot behavior, the techniques developed in this work could be used in other applications in dynamic scenarios, in which the robot has to avoid areas with a high density of moving obstacles (e.g. robotic services in airport or train station halls).

The remainder of the paper is organized as follows. Section 2 discusses the related work. In section 3 the general system concept and the contributions are described. Section 4 presents a general overview of the proposed approach to solving the problem of monitoring the scenario. Section 5 defines the variables and tools used along this work. Section 6 explains the offline planning aspect of our method. Section 7 provides details on the online method used by the robot to select direction and goals, including path planning to avoid dynamic areas. Results from quantitative and qualitative evaluation based on simulations and on real-world experiments, respectively, are presented and discussed in Section 8. Finally, conclusions are detailed in Section 9.

## 2. Related Work

The monitoring of the environment can be approached in various ways. Works such as [1] and [2] propose coverage solutions that involves repeatedly covering the scenario over a specific period, and [3] makes a review of coverage techniques. Patrolling applications are presented in [4], [5], [6], and [7], where robots follow precomputed paths (patterns) cyclically. Works as [8] and [9] develop environment exploration tasks without a preexisting map. Notably, these approaches are designed for static scenarios, and therefore are not longer valid for dynamic ones, because of effectiveness, efficiency and safety. The method developed in this work deals with environment monitoring tasks context. It assumes a prior static map is available, and leverages this information to plan future goals while avoiding some dynamic areas that prevent monitoring and/or lead to potential collisions.

Much of the recent work has primarily focused on navigation among dynamic obstacles. Studies such as [10][11][12][13] develop and evaluate various local planners in human-crowded environments. These methods emphasize robot navigation among pedestrians, assessing several classical and social metrics. However, they do not incorporate a global planner that explicitly accounts for social behaviors when computing and adapting the paths, nor do they anticipate the robot's actions to avoid risky and crowded areas.

To anticipate possible collisions, some works focus on developing global planners that integrate social rules into classical path-planning algorithms. In [14], an RRT* global planner is adapted by computing a global plan based on learning the weights of its cost functions from human trajectories, thereby allowing the planner to mimic demonstrated behaviors. Depending on the learned context-specific weights, the computed path may or may not intrude into human personal space. In [15], the authors modify the classic Elastic Band (EB) approach by incorporating human movement patterns to facilitate human-robot co-navigation. The Fast Marching Method (FMM) is used for social navigation in [16], where an "actuation" space around humans is included to anticipate their movements. The work in [17] employs an optimization approach to generate paths that simulate possible human trajectories. In [18], the authors use A* to reason about the spatial relationship between robots and humans over time, applying a Gaussian-based social cost model to predict human movements. However, a key limitation of all these works is the low number of humans considered in the avoidance process. Additionally, analyzing each person individually can significantly increase computation time, making real-time calculations prohibitively slow.

Some works combine global and local planners. In [19], the A* algorithm uses a heuristic cost derived from the Dynamic Object Velocity-Time Space (DOVTS) to expand nodes that do not lead to potential collisions. Similarly, in [20], the authors develop a navigation system based on an FMM path planner that integrates social costs to avoid small groups of people. However, like the previously mentioned approaches, these methods focus on scenarios where the robot is forced to navigate among humans. In contrast, there are situations where it may be more efficient for the robot to bypass densely occupied areas, if possible.

The objective of our work is to enable the robot to plan and navigate in larger environments containing static obstacles as well as small or large groups of people moving through the space. Depending on the application context, and the potential risks of collision or reduced visibility in crowded, dynamic areas—such as in monitoring tasks—the global plan and motion execution will adapt during each control period. In designing the global plan, we incorporate social principles, as outlined in [10], such as safety, politeness, understanding other agents to reduce potential conflicts before collisions, and contextual appropriateness. Additionally, we consider the concept of people density in different parts of the environment to identify dynamic areas, guiding the decision of whether to enter a populated area or avoid it.

The authors of [21] propose a method to avoid regions populated by obstacles. They develop a Crowd-based Dynamic Blockages (CBDB) layer to identify and evaluate whether to traverse regions with obstacles, based on their density. Our method is similar in concept but with some key differences. Our approach, *MADA*, is less conservative regarding obstacle-populated regions. Instead of considering large regions as in CBDB, the robot defines dynamic areas to be avoided based on the observed movement of obstacles. The path planning approach in [21] requires evaluating several paths to select the best one, whereas *MADA* builds the dynamic areas and computes a single path. Given these similarities, we will compare the performance of CBDB, employed in a monitoring mission, against our developed *MADA* approach.

Additionally, we demonstrate the advantages of integrating dynamic areas with a social path planner in a monitoring mission. We incorporate the FMM-based social path planner from [16] into our proposed method. By including information about dynamic areas, we show that the robot improves mission outcomes in terms of collision reduction and monitoring efficiency, even when using social planners like [16] for monitoring missions.

## 3. System concept and contributions

The system concept is illustrated in the sequence of Fig.2, where the robot monitors the environment, selects the next goal (next region of interest for monitoring), computes a first trajectory towards it, identifies the risky dynamic areas, re-computes a path avoiding them to the next region of interest, iteratively repeating the process

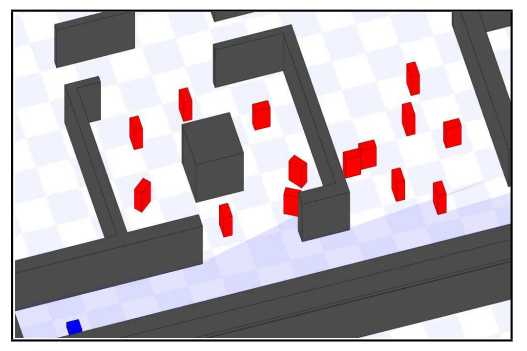

(a) Scenario

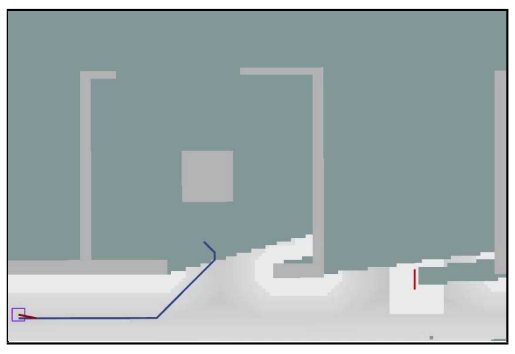

(b) Initial planned path

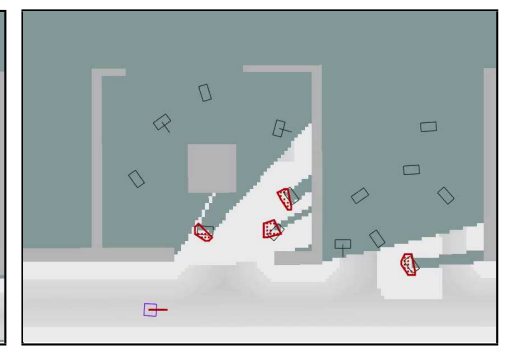

(c) Dynamic areas construction

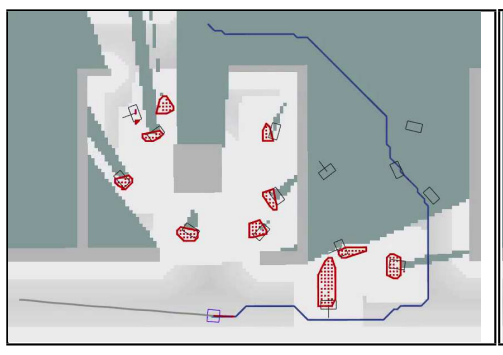

(d) Re-planned path after building dynamic areas

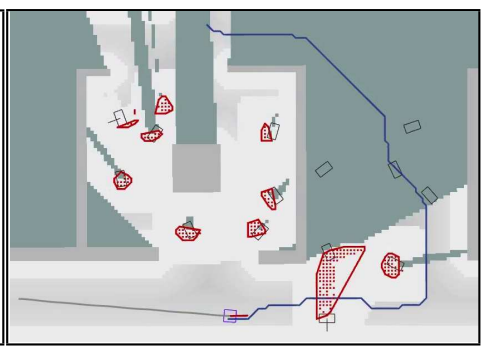

(e) New path through dynamic area

Figure 2: Illustration of the monitoring mission. In Fig.2(a), the robot (depicted in blue) monitors the scenario in the presence of dynamic obstacles (depicted in red). Fig.2(b) illustrates the robot's planned path towards a region to be monitored. The zones observed and not observed by the robot during the motion up to the current location are represented in white and green, respectively. The robot (depicted as a blue square) follows the path (shown as a blue line) to reach the next goal for making new observations. In Fig. 2(c) while following the path, the robot monitors the scenario, observes dynamic obstacles (depicted as black rectangles), and builds dynamic areas (depicted as red polygons). These dynamic areas are constructed based on the observed positions (depicted as red points). If the dynamic are dense enough, the robot obtains a new goal and calculates a new path to reach the goal throughout the observed and the unknown areas, avoiding the risky dynamic areas, as shown in Fig.2(d). Due to the movement of obstacles, the dynamic areas are extended. Consequently, the robot adjusts its path to traverse less occupied dynamic areas, as depicted in Fig.2(e).

until the mission is completed. It's crucial to highlight that only dense dynamic areas, categorized as risky, influence their traversability. Conversely, non-dense dynamic areas do not prevent them from being crossed, allowing the robot to navigate through them while avoiding obstacles that may emerge in these regions. Note that in a monitoring mission, the robot has not to cover all the scenario surface as in a coverage mission. Therefore, the robot only moves into a region up until it is completely observed from a viewpoint, which depends on the visibility range of the sensor used for monitoring, as can be seen in Fig.2. Once that region is observed, the planner computes a new path to the next selected region.

Thus, the contribution of this work this work is the development of a global planner to monitor a scenario, avoiding obstacle-crowded regions. The method follows a two-step approach to plan and execute monitoring missions in dynamic environments, ensuring safety in the context of such missions. In the first step, the *Offline planner* computes an adjacency graph from the prior static map, guiding the process of planning the mission, by connecting the successive areas established as subgoals to be reached. In the second step, an *Online planner*, based on the previous graph, detects every control period the dynamic areas and re-plans the path either to avoid the potentially risky ones, or to traverse them, depending on an occupancy criterion.

The method is evaluated in different scenarios, comparing the results based on metrics with other two techniques, one more conservative regarding the risky areas, and the other greedy, in which there is not an explicit representation of dynamic areas for planning paths taking them into account. An experimental evaluation was performed on a real robot to demonstrate the method's performance in real-world conditions.

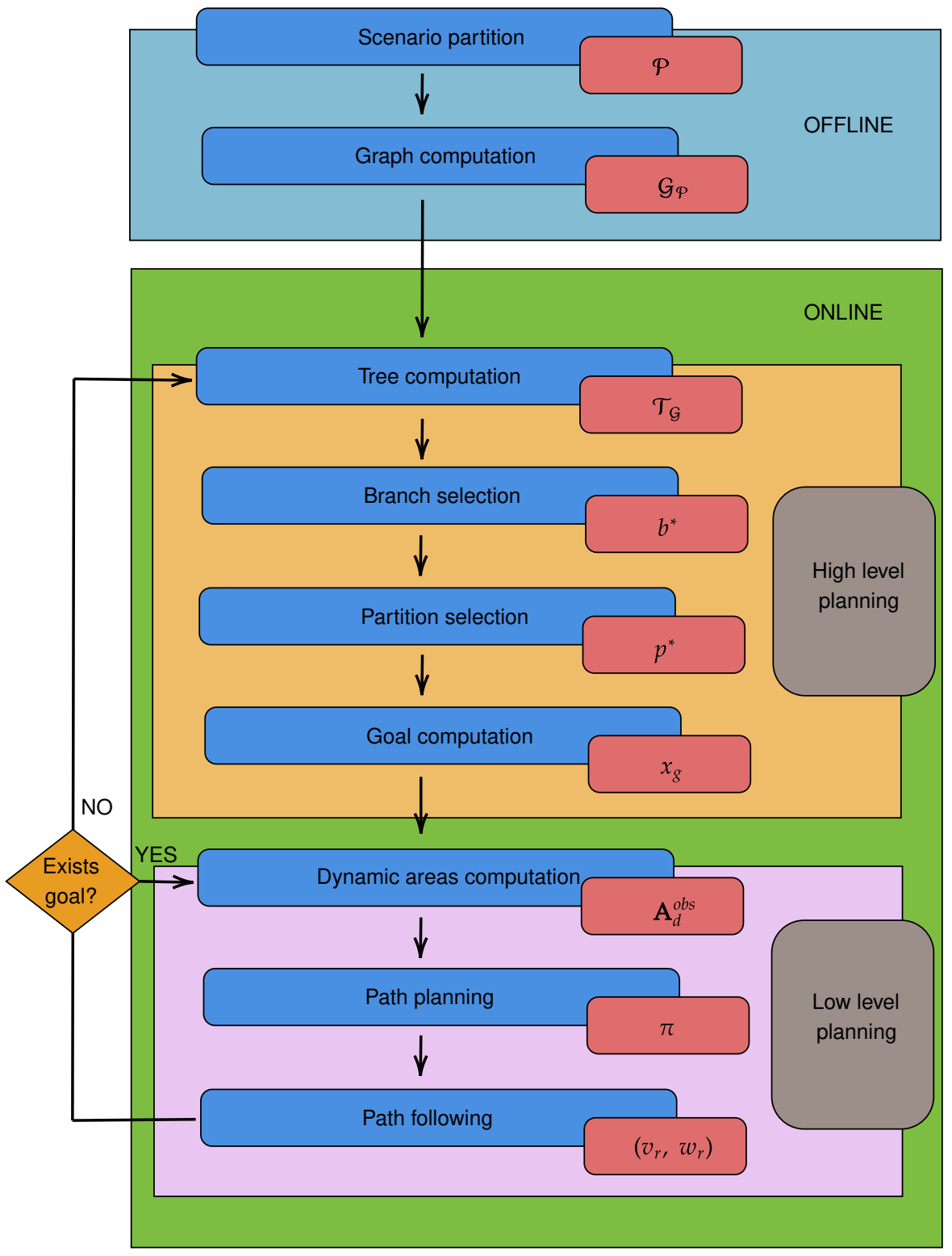

Figure 3: System Overview. The output variables in each module are explained in section 5.1

## 4. System overview

The method computes the global plan and executes the motion in two-steps, as illustrates Fig.3:

- *Offline planning:* From the prior static map, the first step obtains the distribution/layout of the environment, by splitting the scenario into partitions and obtaining the adjacency graph between them, to guide the process of planning the mission. During this phase, the workload cost for monitoring each partition and the time needed to move between partitions are estimated. These estimates are vital for developing an efficient plan to observe the entire scenario during the monitoring mission.
- *Online planning:* Once the mission starts, this planner selects the next region to monitor, based on the estimates of the offline planner, and computes the goal position where the robot will see this region. Then, the planner computes the trajectory towards the new goal. This trajectory avoids risky dynamic areas based on the occupation level of the dynamic obstacles. Once the execution of the initial plan is launched, the online planner recomputes the initial trajectory towards a new goal when a new risky dynamic area is detected, based on the area occupancy, avoiding this area, executing the plan with a path follower.

During the process, the method tries to dynamically compute trajectories that maximize the observed scenario, whilst avoids returning to areas already seen, if it can be achieved. This way, the final trajectory will be as shorter as possible. Figure 4 represents the whole trajectory followed by the robot to complete the mission.

Sections 6 and 7 describe in detail the *Offline planning* and the *Online planning*, respectively. To understand the whole process, the Figures 5, 6, and 7 depict the successive steps followed. In Fig.5, after splitting the scenario into partitions, a graph of the partitions is obtained. This graph will serve as a guide for monitoring the whole scenario. In Fig.6 the steps of the Online high-level planning are depicted, for reaching a first region of interest as a goal (in red) in Fig.6(c) from the robot position in Fig.5(a). The steps of the Online low-level planning are shown in Fig.7, in charge of detecting the dynamic areas to be avoided, and planning the trajectory towards a new goal in a next region of interest. This way the robot simultaneously monitors the new region and is able to inspect the previous avoided one from a new point of view.

## 5. Notation and basic tools

### 5.1. Notation

We consider a differential robot equipped with a 360º field of view laser. The robot monitors a 2D world represented as a grid, defined by the variable $M$, which corresponds to the static map of the scenario. The variable $x$ represents a position on the grid and $\mathbf{x}$ denotes a set of positions on the grid. $M(x) = 0$ when $x$ contains an obstacle and $M(x) = 1$ when it is free space. The positions of the static obstacles are expressed as $\mathbf{x}_{so}$. Employing the static map $M$ with $\mathbf{x}_{so}$, the robot splits the scenario into partitions expressed as $\mathcal{P}$, see Fig.5(b). The adjacency between the partitions are obtained with the graph $\mathcal{G}_{\mathcal{P}}$, see Fig.5(c). In order to choose the next goal to go to make observation, the robot evaluates all the possible directions computing the tree $\mathcal{T}_{\mathcal{G}}$, Fig.6(a). Then, the branch of the minimum cost $b^* \in \mathcal{T}_{\mathcal{G}}$ is selected. Then, the robot chooses the next partition from the branch, denoted as $p^* \in b^*$. The observation of the partition is made from the computed goal, expressed as $x_g$, see Fig.6(c).

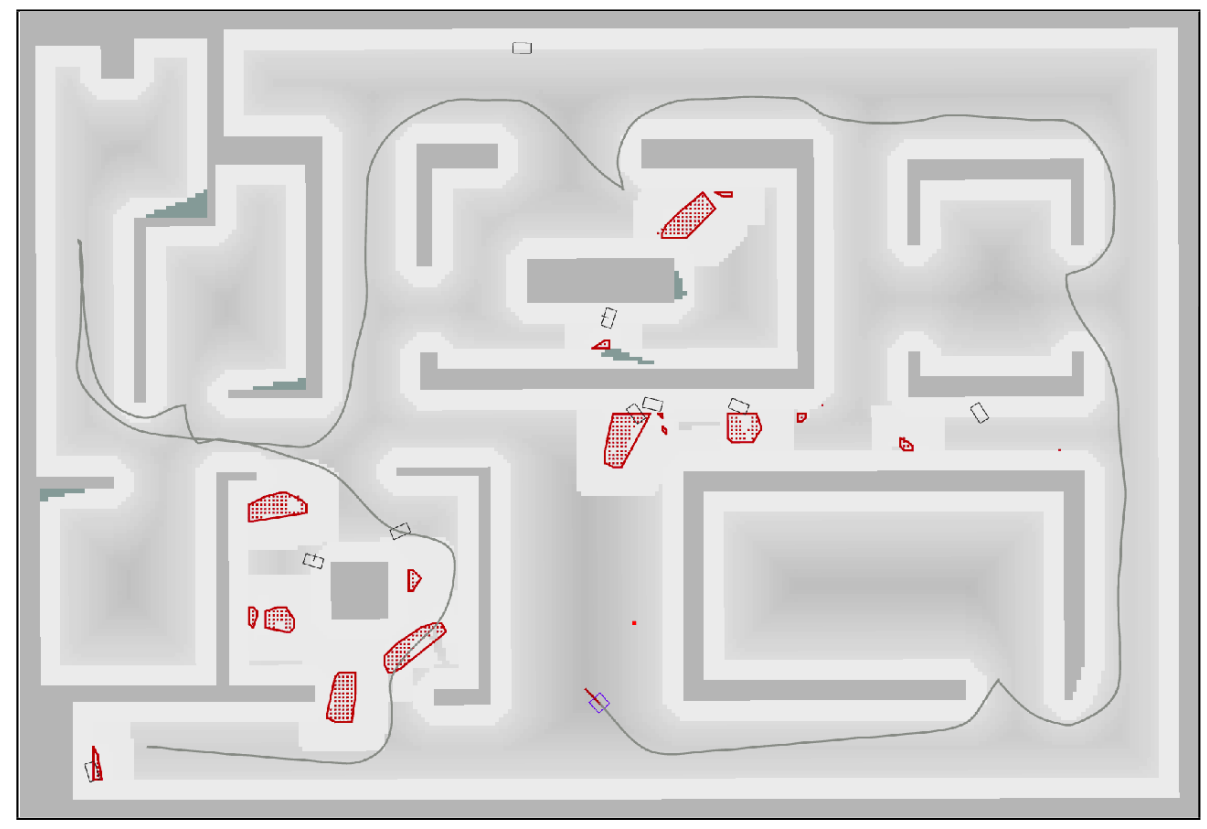
Figure 4: The robot travels along trajectories away from static obstacles during the mission, monitoring in this case the most part of the scenario.

The position of the robot is $x_r$ and its visibility range is $r_v$. All the visible positions from $x_r$ are denoted as $\mathbf{x}_v$. The robot moves in a scenario with dynamic obstacles, whose positions are denoted as $\mathbf{x}_{do}$. When the robot observes the motion of dynamic obstacles, see Fig.7(a), it stores their trajectories $\chi_{do}$ and constructs the dynamic areas, expressed as $\mathbf{A}_d^{obs} = [\mathrm{A}_{d_1}^{obs}, ..., \mathrm{A}_{d_N}^{obs}]$, $\mathbf{x}_{do} \in \mathbf{A}_d^{obs}$, see Fig.7(b). In order to move to $x_g$, the robot computes the path $\pi$, defined as a sequence of adjacent positions, depicted in Fig.7(c). The path $\pi$ will avoid the dynamic areas $\mathbf{A}_d^{obs}$, based on the density of the obstacles within them. So that, if the areas are dense, they are considered risky and will not be traversable. To determine if an area is dense, an occupation threshold $O_{th}$ is applied. To follow the path $\pi$, the robot obtains the next destination $x_{dest} \in \pi$, computing the linear and angular velocities to apply, $(v_r, \omega_r)$.

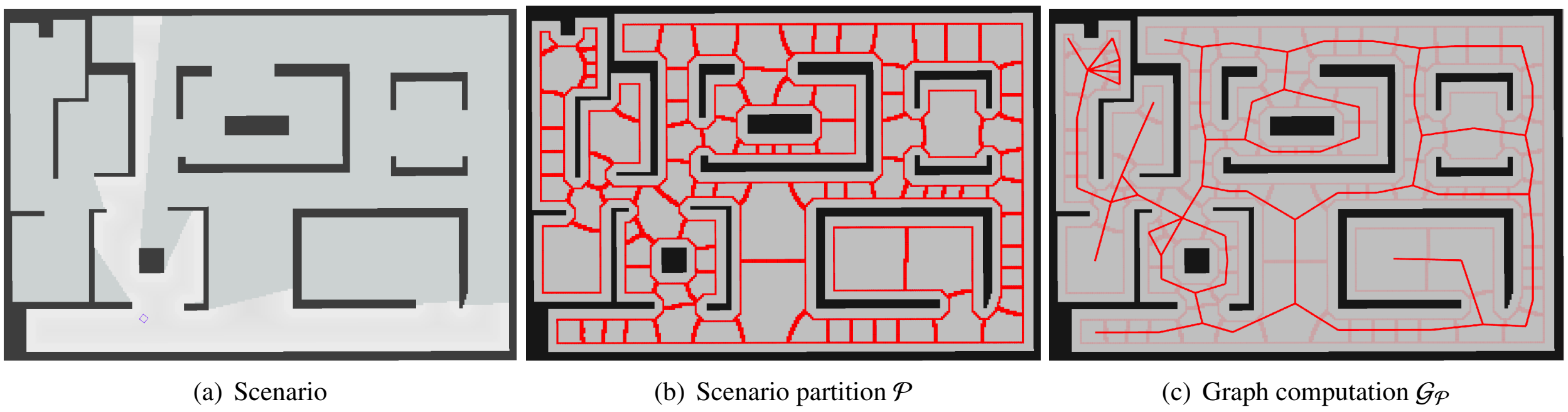

(a) Scenario (b) Scenario partition $\mathcal{P}$ (c) Graph computation $\mathcal{G}_{\mathcal{P}}$

Figure 5: Offline planning to extract information of the scenario shown in Fig.5(a). Fig.5(b) shows the partition of the scenario, that will be sequentially visited and used to estimate the cost to monitor each of them. Fig.5(c) illustrates the graph of the partitions, which is used to know the connections between partitions and to estimate the cost of traveling between them.

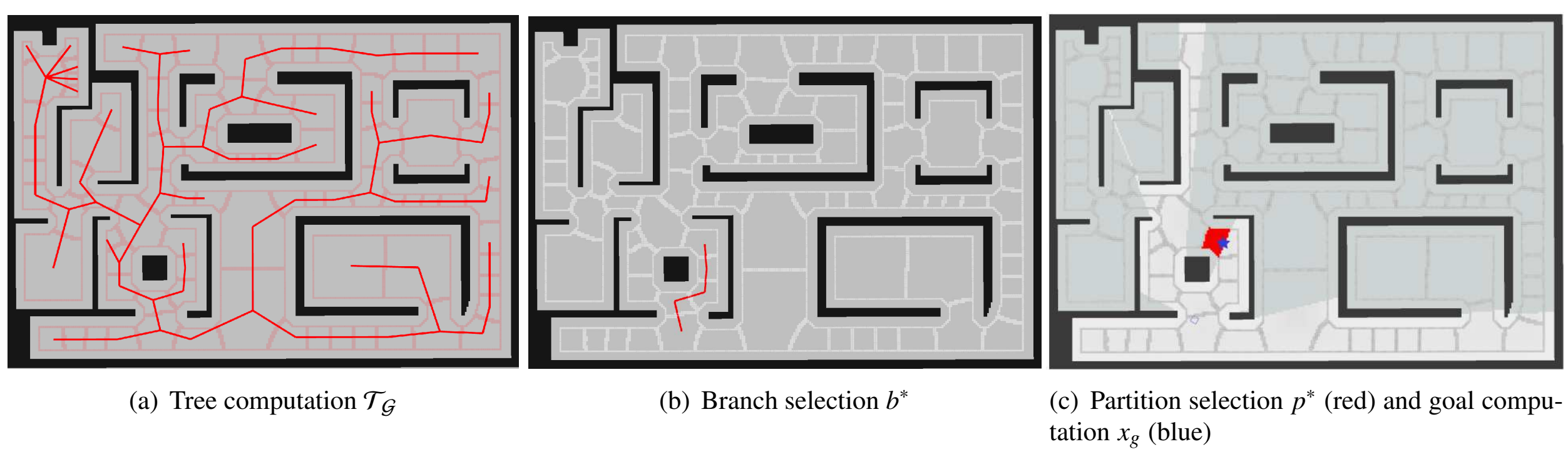

(a) Tree computation $\mathcal{T}_{\mathcal{G}}$ (b) Branch selection $b^*$ (c) Partition selection $p^*$ (red) and goal computation $x_g$ (blue)

Figure 6: Online high-level planning of the proposed approach to select the next goal to make observations. At each moment, when the robot searches a new goal to observe the next set of points, it executes these sequence of methods. Fig.6(a) illustrates the tree obtained for the position of the robot of Fig.5(a). In Fig.6(b), the robot selects the branch of minimum cost from all the possibles of the tree. Fig.6(c) shows the selected partition to be observed (red points) from the computed goal $x_g$ (blue star).

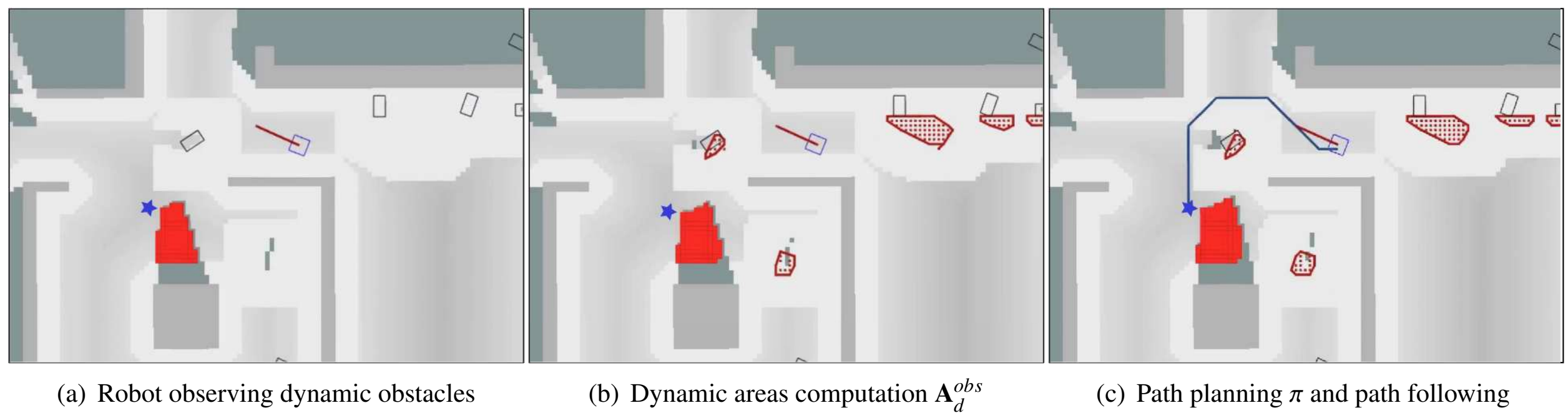

(a) Robot observing dynamic obstacles (b) Dynamic areas computation $\mathbf{A}_d^{obs}$ (c) Path planning $\pi$ and path following

Figure 7: Online low-level planning to avoid dynamic obstacles. Fig.7(a) depicts the robot (blue square) going to the goal (blue star) to observe the points of the partition (red points), in presence of moving obstacles (grey rectangles). In Fig.7(b), the robot builds the dynamic areas (red polygons) from the movement of the obstacles (red points) in successive instants. Fig.7(c) illustrates the obtained path (blue line) to avoid the dynamic area. The red line from the robot center represents the destination to the point on the path to follow it, computing the velocity commands to drive to this destination.

### 5.2. *Fast Marching Method as basic tool*

The Fast Marching Method (FMM) [22] plays a crucial role in various aspects of our proposed monitoring mission approach, offering efficiency and versatility. We leverage this mathematical tool across different stages, benefiting from its unique properties to compute diverse intermediate results. FMM stands out among other path planners for our purposes, such as Probabilistic Roadmaps (PRM) [23] and various versions of Rapidly-exploring Random Trees (RRT) [24, 25, 26, 27], by providing a balance between speed and optimality. Unlike A* path planning method [28], FMM excels by requiring only one gradient computation from the agent's position to all possible goals. This efficiency significantly reduces the computation time, making it a favorable choice for our approach. Moreover, FMM enables the computation of paths that ensure a safe distance from obstacles, contributing to the maximization of the field of view for monitoring, enhancing navigation safety to mitigate collision risks, avoiding to be blocked by moving obstacles near the walls, or increasing the visibility of obstacles in corners, following some social rules described in [10].

FMM works on a grid map, and involves propagating a wavefront from a source location over a surface. The distance from the source is computed at all points (cells) during propagation, using the accumulated distance from the neighbors, ensuring that surrounding obstacles are always accounted for:

$$|\nabla D| F = 1 \tag{1}$$

where $\nabla D$ represents the distance gradient from the source position and $F$ is the velocity of the wavefront.

$F = 0$ when the cell contains an obstacle, and $F = 1$ if the cell is free space. With these velocity values, $F = [0, 1]$, using the FMM for classical path planning causes the wavefront to propagate uniformly in all directions, avoiding obstacles and obtaining the distance gradient. By simply descending this gradient from any point, it will lead to the source of the wavefront, thus obtaining the shortest path to the source point while avoiding obstacles. FMM is used to split the scenario into partitions in the Offline planning, and to compute the trajectories which avoid the dynamic areas during the Online planning.

In the following sections, we will explain how we adapted FMM to take advantage of its properties for the proposed method.

**Algorithm 1** Scenario partition

**Require:** Static map $M$, static obstacles $\mathbf{x}_{so}$

1: $\nabla D_{so} \leftarrow compute_gradient(\mathbf{x}_{so}, M)$
2: $\mathbf{x}_p \leftarrow \emptyset, \nabla D'_{so} = \nabla D_{so}$
3: **while** $any(\nabla D'_{so}) > 0$ **do**
4: $\quad [v_p, x_p] \leftarrow find_max(\nabla D'_{so})$
5: $\quad \mathbf{x}_p \leftarrow x_p$
6: $\quad \nabla D'_{so}(x_p - v_p, x_p + v_p) = 0$
7: **end while**
8: $\nabla D_c \leftarrow compute_gradient(\mathbf{x}_p, \nabla D_{so})$
9: $\mathcal{P} \leftarrow obtain_partitions(\nabla D_c)$
10: **return** $\mathcal{P}$

## 6. Offline planning

Before the mission begins, the robot extracts offline the information of the environment in which is moving. In this section we describe the offline planning procedure. The division of the scenario in partitions is explained in Section 6.1. Section 6.2 details the computation of the adjacency graph from the obtained partitions.

### 6.1. *Scenario partition*

Several studies have addressed scenario partitioning, such as [29] for room identification and [30] for robotic room-by-room clearance. However, for our monitoring-focused problem, partition shapes are not critical. The robot only needs information on partition size for observation times and connections between partitions to select the next one, discarding those potentially unreachable due to dynamic obstacles. Our approach employs the Fast Marching Method (FMM) for scenario partitioning, incorporating static obstacles and using interpolated distances to propagate the wavefront between the points of the partitions, differentiating from traditional Euclidean distances employed in Voronoi tesselation [31].

Algorithm 1 describes the method for splitting the environment with FMM, as illustrated in Fig.8. The gradient from the static obstacles, $\nabla D_{so}$, is computed from the map of the scenario $M$. The wavefronts are initialized from the static obstacles $\mathbf{x}_{so}$. The maximum value of the gradient, $v_p$, is searched iteratively and is placed at the partition point $x_p$. This position will be the furthest point from the obstacles within the partition and will be a source of the wavefront of FMM. The values of the gradient within the distance $v_p$ are set to zero (equivalent to an obstacle), to avoid selecting several wavefront sources in the same space, Fig.8(a). Then, the gradient $\nabla D_p$, Fig.8(b), is obtained by propagating the wavefront from all the partition points $x_p \in \mathbf{x}_p$ using

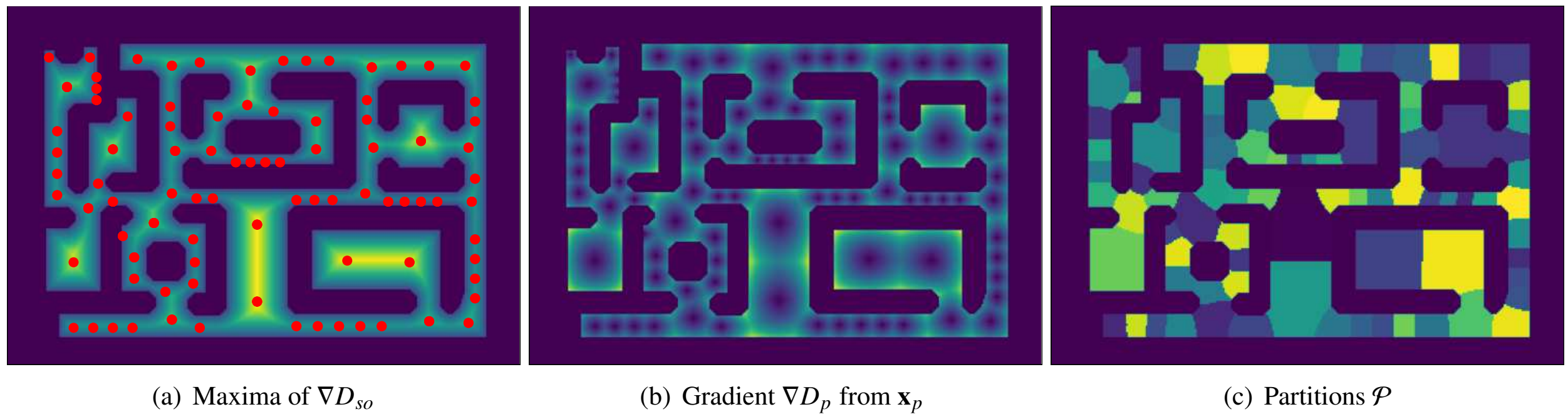

(a) Maxima of $\nabla D_{so}$ (b) Gradient $\nabla D_p$ from $\mathbf{x}_p$ (c) Partitions $\mathcal{P}$

Figure 8: Environment partition procedure. In Fig.8(a), the maxima of $\nabla D_{so}$ are found. These maxima correspond to the point of maximum value of the gradient in each partition of the environment $\mathbf{x}_p$. Fig.8(b) illustrates the gradient $\nabla D_p$, obtained by propagating the wavefronts from $\mathbf{x}_p$. The partitions of the scenario are extracted from $\nabla D_p$ and depicted in Fig.8(c).

the distance transform $\nabla D_{so}$ as the velocity of propagation. This ensures that the wavefront propagates faster in wide spaces, such as rooms or corridors, slowing down and colliding in narrow spaces, such as doors, Fig.8(b). Finally, the partitions are obtained as the regions covered by each of the wavefronts, coming from $\mathbf{x}_p$, as illustrated in Fig.8(c). As illustrated in the figure, the resulting partitions have a polygonal shape and a notable feature is that all straight lines connecting each point on the polygon edge with $x_p$ lie within the polygon. Therefore, if the visibility range of the sensor $r_v$ is large enough, the robot moves towards $x_p$ to observe the entire partition, not needing to move to other positions inside the region for this observation, thus aiming to reduce the total mission time.

### 6.2. *Graph computation*

The graph of the environment $\mathcal{G}_\mathcal{P}$ is obtained from the adjacency between the different partitions described in the previous subsection. Figure 9 illustrates the adjacency graph of the partitions obtained in Fig.8(c). This ensures that the robot always has the information about the distribution of the different partitions in the environment, in order to properly choose the next one to monitor. This graph serves to estimate the cost of observing the partitions and travelling between them, to guide the monitoring of the whole scenario.

As can be seen in Fig.9, multiple nodes or edges are located in the same regions, such as corridors or rooms. This occurs because the nodes' positions are adjusted based on the available free space, i.e., the distance to nearby obstacles. This approach has two main advantages. First, it ensures that the robot can monitor the entire partition (assuming sufficient view range), which wouldn't be guaranteed by simply traversing the edge if nodes were only at the beginning and at the end of the corridor. Second, in dynamic environments, certain regions may become non-traversable or need to be avoided. Hence, it is more efficient to consider smaller, adjusted partitions rather than large areas. For example, in an elongated corridor, it is better to avoid only the sections occupied by dynamic obstacles, rather than discarding the entire corridor.

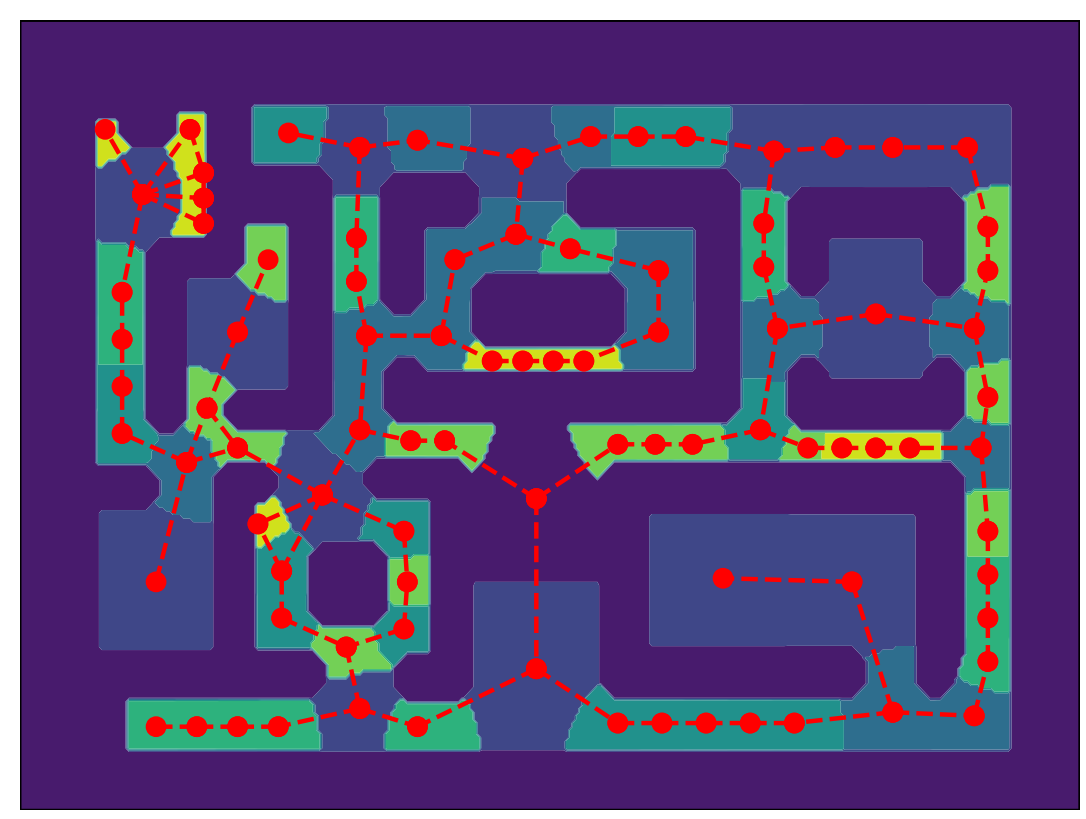

Figure 9: Graph $\mathcal{G}_\mathcal{P}$ for partitions of Fig.8(c).

## 7. Online planning

As represented in the diagram of Fig.3, once the mission starts, the robot selects new goals and moves to them avoiding the dynamic obstacles. High-level planning selects the partition $p^*$ from $\mathcal{G}_\mathcal{P}$ obtained in the Offline planning, computing the new goal $x_g$ to visualize the partition. The low-level planning detects the dynamic areas $\mathbf{A}_d^{obs}$. If they exist, a new path avoiding them is computed and executed, with the robot following it by applying the computed velocity $(v_r, \omega_r)$. Both planners are explained in the next Sections 7.1 and 7.2.

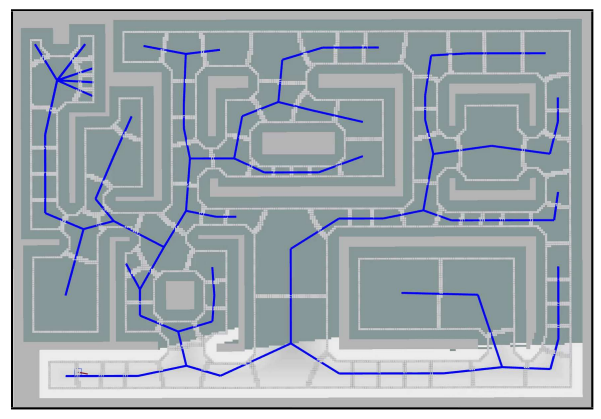

(a) Example of tree $\mathcal{T}_{\mathcal{G}}$

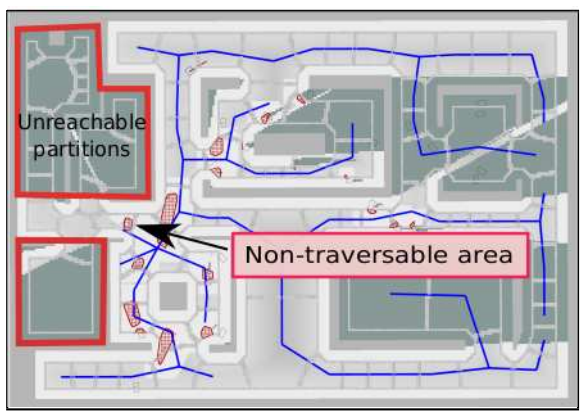

(b) Unreachable partitions due to obstacles

Figure 10: Computation of the tree graph $\mathcal{T}_{\mathcal{G}}$.

### 7.1. High-level planning

The next goal to be reached in the next region of interest is calculated by applying in sequence four procedures: tree computation, branch selection, partition selection and goal computation. They are explained in the following subsections.

#### 7.1.1. Tree computation

Firstly, the robot has to know the direction where to move to continue monitoring the scenario. Since, the robot has the information of the scenario to monitor, that is the partitions $\mathcal{P}$ and the graph $\mathcal{G}_{\mathcal{P}}$, it must select a sequence of edges of the graph (branch) in order to know the direction. For this purpose, we compute the tree from the graph, denoted as $\mathcal{T}_{\mathcal{G}}$, Fig.10. Its root or origin is the partition where the robot currently is and the edges are selected executing Dijkstra's algorithm [32], obtaining the connections between the partitions. To execute Dijkstra, we consider the cost to travel between partitions, computed as $c_t = d_{ij}/v_{max}$, where is $d_{ij}$ is the Euclidean distance between the vertices of partitions $i$ and $j$ and $v_{max}$ is the maximum robot linear velocity. This way, we obtain the tree with all the possible paths to reach all the partitions, traveling between the partitions, as illustrated in Fig.10(a). This ensures that the path formed by each branch of the tree represents the path of minimal cost from the robot's partition to every partition in the scenario.

The knowledge of the possible branches is also important for our problem since there are dynamic obstacles moving in the scenario. Some of the tree branches might be unreachable. For instance, in Fig.10(b) some partitions are discarded to be reached because the paths to them in the tree are blocked by the moving obstacle at the left area in the figure. This information is included in the Low level planner, as will be described in Section 7.2.2.

#### 7.1.2. Branch selection

Due to the dynamism in the environment, it is not possible to know the best direction to take, as it may not be reachable. Therefore, we propose to use the *Greedy* approach to select the branch, minimizing the estimated cost of traversing $c_t(b_i)$ and monitoring $c_m(b_i)$ the partitions of branch $b_i$,

$$b^* = \underset{b_i \in \mathcal{T}_{\mathcal{G}}}{\operatorname{argmin}}(c_t(b_i) + c_m(b_i)) \tag{2}$$

where,

$$c_t(b_i) = dist(\pi_{b_i})/v_{max} \tag{3}$$

with $\pi_{b_i}$ denoting the path formed by the sequence of partition points $x_p$ of the partitions of branch $b_i$ (that is, the sequence of edges), and $dist()$ expresses the function to sum the Euclidean distances between $x_p$ of $\pi_{b_i}$.

The estimated monitoring cost $c_m(b_j)$ in the partition $\mathcal{P}_j$, is calculated with the following expression, which depends on the visibility range $r_v$:

$$c_m(b_j) = \begin{cases} 1, & r_v \geq d_{max_j} \\ r_v \frac{|\mathbf{x}_{ns_j}|}{|\mathbf{x}_{t_j}|}\left(\frac{d_{min_i}}{r_v} \cdot \frac{d_{max_j}}{r_v} - 1\right), & r_v < d_{max_j} \end{cases} \tag{4}$$

where $\mathbf{x}_{ns_j}$ denotes the points not yet seen in partition $\mathcal{P}_j$, $\mathbf{x}_{t_j}$ are all the points of $\mathcal{P}_j$, $d_{min_j}$ and $d_{max_j}$ are Euclidean distances from partition points $x_p \in \mathcal{P}_j$ to the closest and the farthest point of the partition, respectively. If the visibility range of the sensor $r_v$ is less than $d_{max_j}$, the robot will have to traverse the partition to observe all points. Finally, the total cost of monitoring the branch is computed from the summation of the costs of the partitions included in the branch, $c_m(b_i) = \sum_{b_j \in \mathcal{P}_j} c_m(b_j)$.

#### 7.1.3. Partition selection

Once the branch $b^*$ is selected, it is necessary to know the space to be monitored. It is chosen as the first partition $p^*$ of the branch with points that have not yet been seen, as illustrated in Fig.11.

#### 7.1.4. Goal computation

After selecting the next partition to monitor, the robot selects the goal position to move to, see Fig.12. This procedure is detailed with Alg.2. When the sensor range is larger than the distance from the $x_p$ to the farthest point of the partition, the robot is able to visualize absolutely the entire partition, due to the way of constructing the partitions, explained in Section 6.1. The partition point $x_p$ is selected as the goal. Otherwise, all the unseen points of the partition are grouped by adjacency,

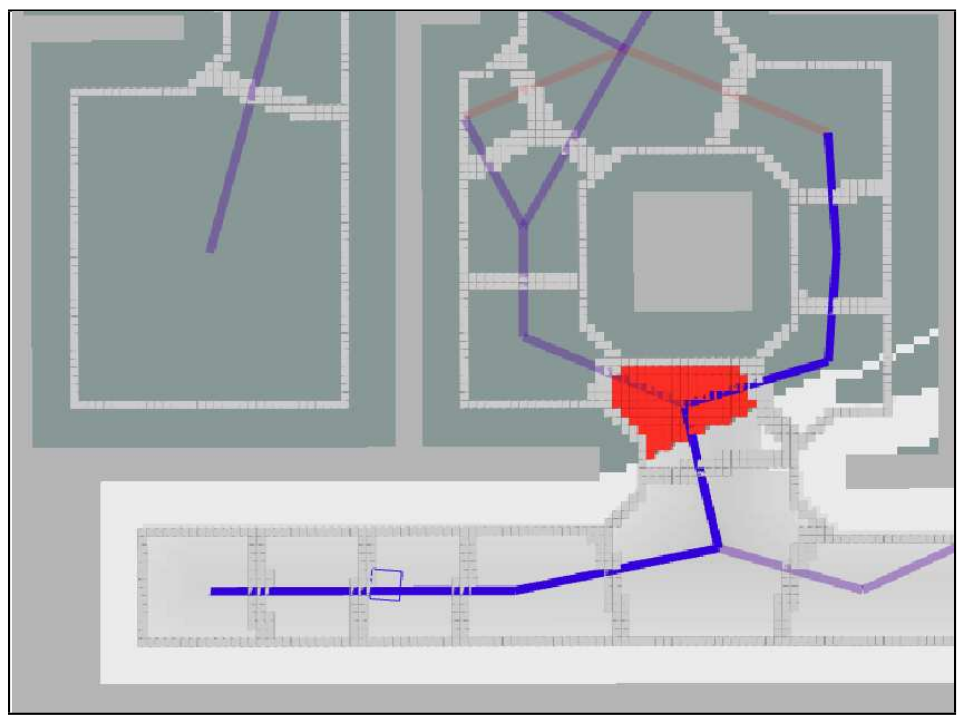

Figure 11: Selection of the partition of a branch represented in blue. The robot chooses the partition that still has points to be observed, shown in red. The gray colour in the partitions indicates observed points, while the green colour indicates unobserved points.

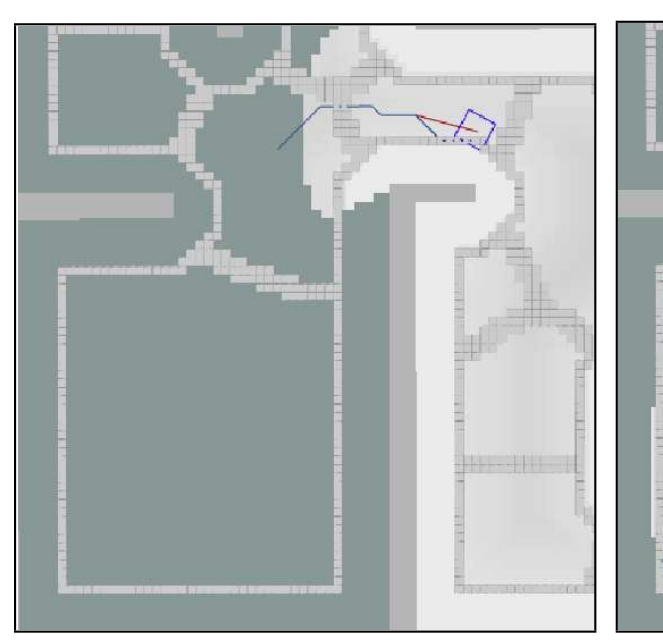

(a) $v_r \geq d_{max_i}$ so the goal is the point $x_p$ of the partition $i$

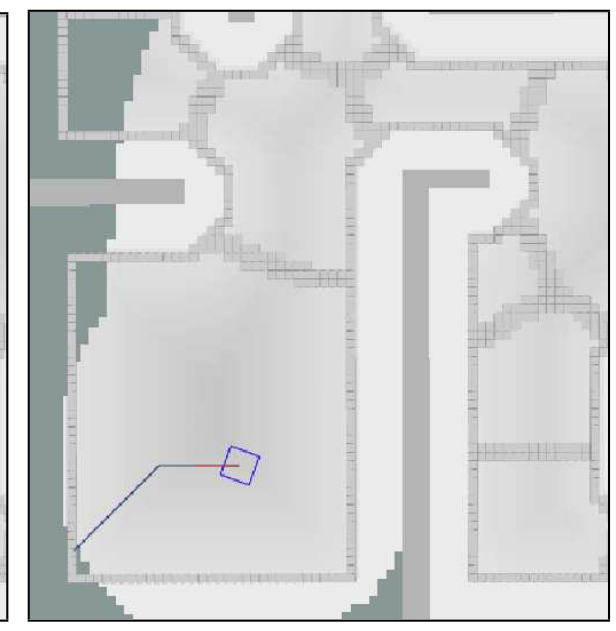

(b) The goal is selected as the closest point of non-seen points, in green

Figure 12: Goal selection. In Fig.12(a), the visibility range is greater than the maximum distance of the partition, so all the points are observed from $x_p$. In Fig.12(b), the goal is the closest point from the smallest set of points in the partition to be monitored.

**Algorithm 2** Goal computation

1: **if** $r_v \geq d_{max_i}$ **then**
2: $\quad x_g = x_{p_i}$
3: **else**
4: $\quad [\mathbf{x}_1, ..., \mathbf{x}_k] \leftarrow group_points(\mathbf{x}_{ns_i})$
5: $\quad \mathbf{x}^* = \{\mathbf{x}_j \mid |\mathbf{x}_j| = min([|\mathbf{x}_1|, ..., |\mathbf{x}_k|])\}$
6: $\quad x_g = \{x \mid min(\nabla D_r(x)),\ x \in \mathbf{x}^*\}$
7: **end if**
8: **return** $x_g$

selecting the one containing the least number of points. The value $\nabla D_r(x)$ is the gradient from the position of the robot. The minimum value of $\nabla D_r(x)$ allows to select the next goal inside the partition, $x_g$. Whilst the robot moves, the set of unseen points is updated and the routine is re-launched, again selecting the goal as well as the path to it.

When the robot goes to $x_p$, it is expected to view the partition points directly from there. As soon as the robot reaches the target and does not see those points due to moving obstacles, it remains in standby. Since the movement pattern of dynamic obstacles is unknown, it is not possible to know when they will move. Therefore, once it reaches the goal, a waiting time $t_{wait}$ is set before discarding the occluded points and going to a new goal. This time is estimated as:

$$t_{wait} = 2\, d_{max_i} / v_{max} \tag{5}$$

where $2\, d_{max_i}$ represents the greatest distance the robot must travel to traverse the partition $\mathcal{P}_i$.

### 7.2. *Low level planning*

During the mission, the robot must effectively track and avoid moving obstacles like humans or robots. Existing human-tracking methods [33, 34, 35] are computationally demanding, posing a challenge for our goal of swift path planning and collision avoidance. We propose identifying positions of dynamic obstacles, regardless of their nature, to construct pre-defined dynamic areas for efficient, proactive avoidance. Local planners for dynamic scenarios like VO [36], ICS [37], and DOVTS [38] could be used for this objective, but in the context of a monitoring application, they do not solve the problem of keeping the maximum visibility whilst possible, increasing the risk of collision in populated areas. Furthermore, a global planner is always needed for accomplishing the missions. Therefore, we propose a new low level planner to manage dynamic scenarios, because the classical global planners for static environments are no longer valid. Our proposed approach allows the robot to follow a path with a wide field of view and avoiding dynamic areas to reduce collision risks.

This section explains how this plan is executed in the presence of dynamic obstacles. In Sect. 7.2.1, we explain how dynamic areas are constructed from the movement of obstacles. Path planning with FMM is detailed in Sect. 7.2.2. And finally, the reactive path following method is explained in Sect. 7.2.3.

#### 7.2.1. *Dynamic areas construction*

The procedure of dynamic areas construction is described in Alg.3. The robot detects and updates every control period the moving obstacles $\mathbf{x}_{do}$, computing the dynamic areas $\mathbf{A}_d^{obs}$ from their trajectories $\chi_{do}$, as illustrated in Fig.13. As it will be explained in section 7.2.2, not all the dynamic areas must be avoided, only those having a high density of obstacles that prevent the mobility inside that region, based on a established occupation threshold $O_{th}$.

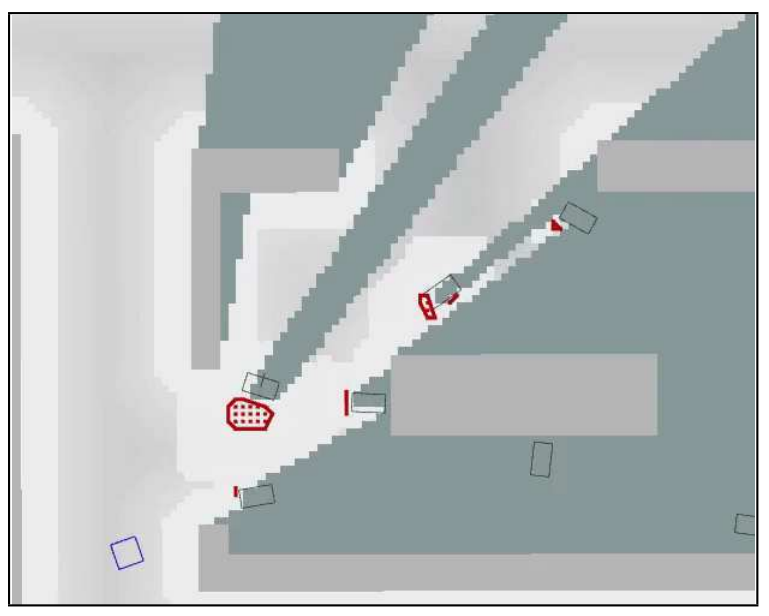

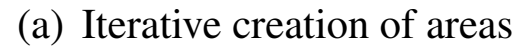
(a) Iterative creation of areas

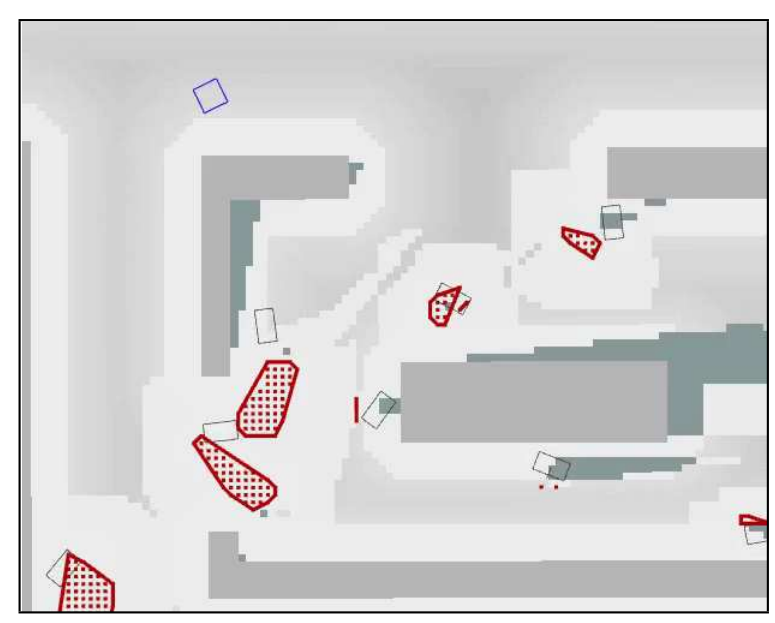
(b) The areas increase with the movement of obstacles

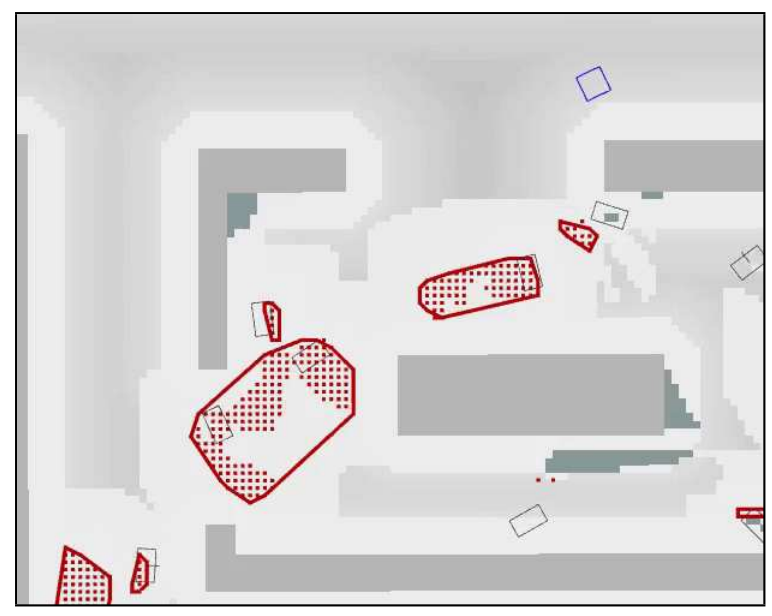
(c) Areas merging after some iterations

Figure 13: Construction of the dynamic areas $\mathbf{A}_d^{obs}$ from the trajectories $\chi_{do}$ traveled by the dynamic obstacles seen by the robot.

**Algorithm 3** Dynamic areas update

1: $\mathbf{x}_{vo} = \{x | x \in \mathbf{x}_v \ , \ M(x) = 0 \ , \ x \notin \mathbf{x}_{so}\}$ ▷ Visible obstacles
2: $\mathbf{x}_{do} \leftarrow update_positions(\mathbf{x}_{vo})$ ▷ Update obstacles list
3: $\chi_{do} \leftarrow \mathbf{x}_{do}$ ▷ Insert in trajectories of obstacles
4: $\mathbf{A}_d^{obs} \leftarrow \emptyset$ ▷ Initialize dynamic areas
5: **for each** $\chi_{do_i} \in \chi_{do}$ **do**
6: $\quad A_{d_i}^{obs} \leftarrow convex_hull(\chi_{o_i})$
7: **end for**
8: **return** $\mathbf{A}_d^{obs}$

*7.2.2. Path planning*

In a monitoring mission, the robot would have to move through areas where its visibility field is expanded. It can be achieved if the planned paths traverse areas away from obstacles. In [39], the authors propose using FMM to keep UAVs far from obstacles; a similar approach is adopted here. To do this, the wavefront is initialized setting the static obstacles as source points, and propagated from the static obstacles through the free space, computing the distance gradient from the static obstacles $\nabla D_{so}$. Setting the velocity map $F = \nabla D_{so}$, propagating the wavefront from a goal position will produce a gradient with higher values in points that are close to obstacles and lower values in points that are far from obstacles. Descending this gradient, the resulting path will traverse points that are far away from the obstacles. Thus, by following these paths, the robot has a lower probability of getting stuck by dynamic obstacles and nearby static ones, leaving clearer spaces for moving.

Another reason for applying such kind of global planning is complying with the named social navigation scenarios as *intersection* and *blind corner*, (as defined in [10]). Planning paths as far as possible from the static obstacles, allows to increase visibility in those scenarios, anticipating avoiding or maneuvers inside of the group.

At the same time, the information of dynamic areas must be included in the velocity map $F$, in order to avoid traversing dense areas. The velocity map, used to propagate the wavefront computing the gradient, is defined using the following expression:

$$F(x) \begin{cases} 0, & x \in \{\mathbf{x}_{so}, \mathbf{x}_{do}\} \\ 0, & x \in A_{d_i}^{obs}, n_{do_i}/|A_{d_i}^{obs}| > O_{th} \\ n_{do}^{max} - n_{do_i}, & x \in A_{d_i}^{obs}, n_{do_i}/|A_{d_i}^{obs}| \leq O_{th} \\ \nabla D_{so}(x) + n_{do}^{max} + 1, & x \notin \mathbf{A}_d^{obs}, x \notin \mathbf{x}_{so} \end{cases} \tag{6}$$

Dynamic obstacle density, denoted as $n_{do_i}/|A_{d_i}^{obs}|$, is computed by considering the ratio of the number of points occupied by obstacles within a dynamic area $i$ ($n_{do_i}$) to the total number of points in this area ($|A_{d_i}^{obs}|$). In this context, the function $F(x)$ is assigned a value of 0 in positions where static or dynamic obstacles are present, as well as in those within dense dynamic areas defined by the $O_{th}$ threshold. It takes non-zero values in positions corresponding to non-dense areas or free space, with a priority for the latter. As follows from eq.6, static and dynamic obstacles are treated similarly in terms of avoidance. However, dynamic obstacles also contribute to defining dynamic areas, helping to avoid crowded regions.

From $F$ values, the robot computes the gradient, as seen in Fig.14. The dynamic areas of Fig.14(a) produce higher values of gradient $\nabla D$ in Fig.14(b). Thus, descending the gradient, these areas have a lower priority compared to free space, obtaining a path $\pi$ surrounding them.

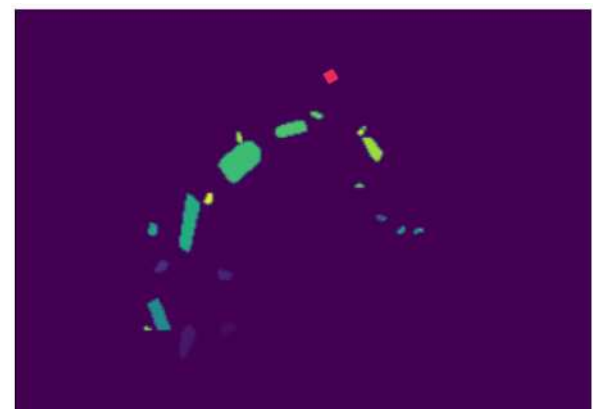

(a) Velocity map $F$ in presence of dynamic areas

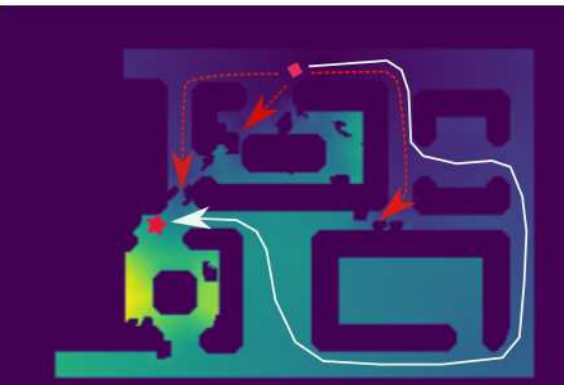

(b) Gradient obtained using $F$

Figure 14: Inclusion of the dynamic areas of Fig.13(c) in the path planning. The velocities map $F$ for the wavefront propagation is depicted in Fig.14(a). The computed gradient $\nabla D$ used by the robot for navigation is illustrated in Fig.14(b). Positions with moving obstacles are removed from the gradient, making them inaccessible. Cells corresponding to dense dynamic areas have higher gradient values and will be surrounded to avoid the risk of a possible collision. The robot (red square), when navigating towards the goal (red star), prevents finding paths (red dashed lines) through dense dynamic areas. Instead, it will circumvent these dynamic areas (white line), covering a larger distance but evading risky spaces.

### 7.2.3. *Path following*

A simple path follower has been developed, which acts as a reactive navigation method, simultaneously following the path, computing velocities compliant with the kinodynamic constraints, and avoiding the surrounding static and moving obstacles. It is not the a central piece in this work and it could be substituted by other reactive planners. It proceeds in two steps: selecting the destination point $x_{dest}$ with Alg.4 and adjusting the velocities to reach this destination by means of Alg.5.

The next subgoal destination is selected in Alg. 4 on the global path $\pi$, considering a larger or shorter distance from the current robot position as a function of its distance to the closest obstacle. Figure 15 outlines the idea and represents the variables involved in the procedure. The value of $d_{\pi-so}$ is derived directly from the gradient of $D_{so}$ at the positions of the path. The algorithm iterates while the distance between the robot and a path point, denoted by $d_{r-\pi}(x)$, is not greater than the distance from this point to the closest obstacle. In the case that dynamic obstacles are detected, a safety distance of $2R_r$ is incorporated to enhance the safety of the robot. This safety distance affords the robot additional time to maneuver and avoid collisions, in those situations where a dynamic obstacle is approaching.

In the second step, Alg.5 obtains the linear and angular velocities $(v_r, \omega_r)$ compatible with the kinodynamic constraints to align to $x_{dest}$, reducing the linear velocity near the obstacles, being $(v_{max}, \omega_{max})$ the maximum velocities of the robot.

**Algorithm 4** Destination selection

1: $d_{\pi-o}^{max} = max(d_{\pi-so}, d_{\pi-do})$
2: **while** $d_{r-\pi}(x) \leq d_{\pi-so}(x)$ **and** $d_{r-\pi}(x) \leq d_{\pi-do}(x) - 2R_r$ **do**
3: ▷ $x$: current position $x \in \pi$; $d_{r-\pi}(x) = \|x_r - x\|$
4: $x_{dest} = x$
5: $d_{dest-o} = min\left(d_{\pi-so}(x), d_{\pi-do}(x)\right) / d_{\pi-o}^{max}$
6: **end while**
7: **return** $x_{dest}, d_{dest-o}$

**Algorithm 5** Velocities computation procedure

1: $\Delta\theta = atan2(x_{dest},\ x_r) - \theta_r$
2: $\omega = \omega_{max} * \Delta\theta$
3: $$\omega^* = \begin{cases} max(\omega,\ \omega_r - |\Delta\omega|), & \omega < 0 \\ min(\omega,\ \omega_r + |\Delta\omega|), & \omega \geq 0 \end{cases}$$
4: **if** $\Delta\theta \neq 0$ **then**
5: $$v = \begin{cases} v_{max} * d_{dest-o} * \left(1 - \frac{|\omega^*|}{\omega_{max}}\right), & |\Delta\theta| < \pi/2 \\ 0, & |\Delta\theta| \geq \pi/2 \end{cases}$$
6: $v_r^* = min(v,\ v_r + \Delta v)$
7: **else**
8: $v_r^* = max(v_{max},\ v_r + \Delta v)$
9: **end if**
10: **return** $(v^*, \omega^*)$

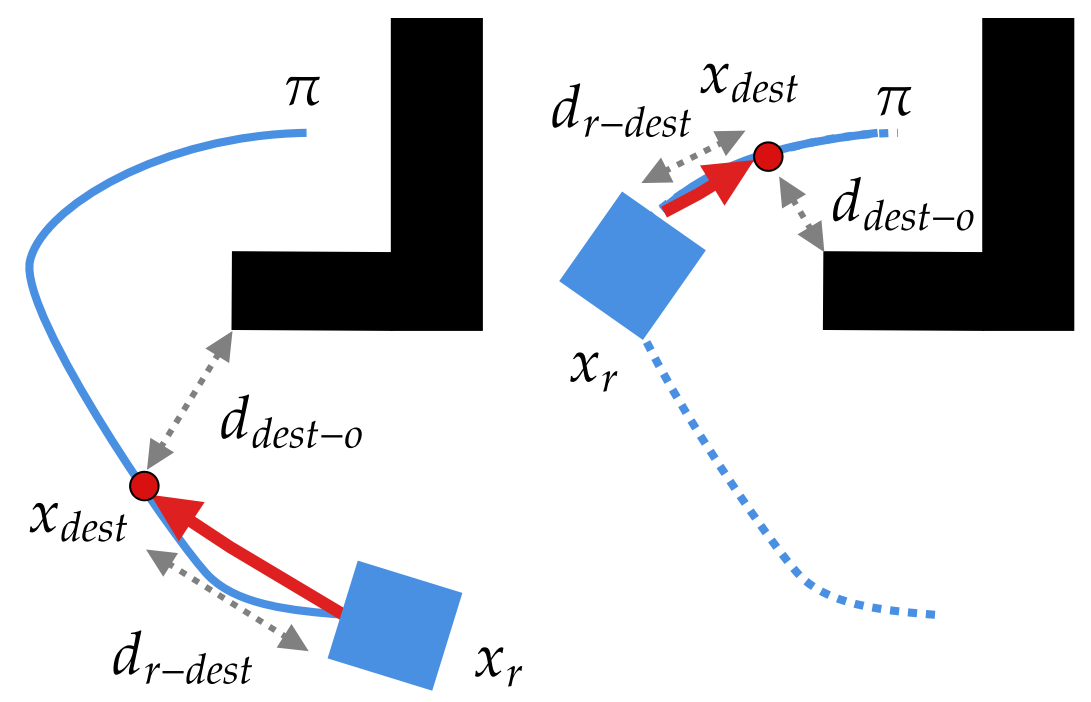

(a) Farthest destination $x_{dest}$ (b) Closer destination $x_{dest}$

Figure 15: Two potential destination selections are presented to follow a path $\pi$, situated in proximity to a static obstacle, depicted with black. In Fig.15(a), the robot $x_r$ selects a destination $x_{dest}$ at the greatest distance $d_{r-dest}$ and the highest velocity $v^*$, based on the distance to the obstacle $d_{dest-o}$, with a greater deviation from the path. Fig.15(b) illustrates the robot selecting a destination at a closer proximity with a lower velocity, in order to align more closely with the path.

## 8. Results

This section presents the results of the proposed approach evaluation. The details of the simulations are defined in Section 8.1. The results of the simulations and their discussion are presented in 8.2. Some extensions of the basic method are described in Section 8.3. Comparisons with other methods are presented in Section 8.4. The validation of the proposed approach with experiments in a real robot is carried out in Section 8.5. The videos of the simulations and real-world experiments are available in the link [1].

### 8.1. Simulation details

In order to provide an exhaustive evaluation of the proposed method, and to highlight one of its utilities, namely its modularity, we consider different aspects:

- *Methods of monitoring the scenario*: strategies, occupancy threshold, safety distance from the obstacles, handling of dynamic areas.
- *Environment-dependent aspects*: scenario, number of dynamic obstacles, velocity of the dynamic obstacles.
- *Extensions of the approach*: projecting obstacle movements, use of other planners.

#### 8.1.1. Scenarios

The proposed approach was tested in two distinct indoor scenarios, depicted in Fig.16. These scenarios differ in their structural characteristics, enforcing different kind of obstacle movements. The first scenario, depicted in Fig.16(a), is smaller in scale, featuring shorter corridors and some wide rooms, with $N_{do} = 10$ and $N_{do} = 25$ moving obstacles, representing non-dense and dense environments, respectively, in which concentrate the obstacles. The second scenario, depicted in Fig.16(b), is larger in scale, encompassing a narrow corridor and clearly defined wider spaces, with $N_{do} = 20$ and $N_{do} = 40$ representing non-dense and dense environments.

The visibility range was set to 30 m in the first scenario and 10 m in the second scenario, with a 360° field of view.

#### 8.1.2. Moving obstacles

In the kind of scenarios considered in simulations and experiments, the moving obstacles are humans, although they could be other vehicles. We have defined four movement patterns and their time intervals to have a similarity with real behaviours: standing still (5-15 sec.), moving in a straight line (5-20sec.), turning right, and turning left (1-3sec. for both). The randomly generated maneuver is checked for potential collisions in the subsequent time step. It should be noted that static obstacles are always guaranteed to be avoided, but collisions between moving obstacles may occur, as no method of reciprocal avoidance is employed.

[1] https://bitly.cx/1Zuv

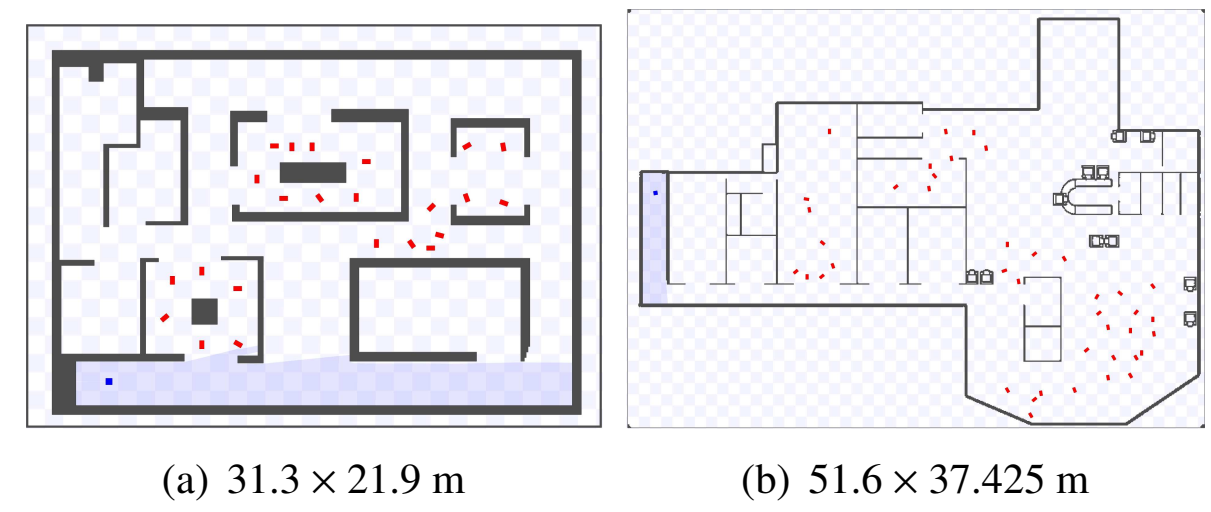

(a) $31.3 \times 21.9$ m (b) $51.6 \times 37.425$ m

Figure 16: Two tested scenarios. The blue square and red rectangles depict the initial positions of the robot and obstacles, respectively.

Many of the obstacles move in almost the same areas, such as humans in a factory, and some obstacles manage to evade the area in which they were initially placed. These areas constitute the so called here dynamic areas, which will be avoided by the global planner for preventing possible disturbance to humans, reduce collision risk, or enforces to surround an area from which the monitoring cannot be correctly achieved due to the presence of people.

We have employed two velocities for the dynamic obstacles: $(v_r/3, \omega_r/2)$ to simulate slow and peaceful obstacle motions and $(v_r, \omega_r)$ for fast and aggressive motions.

#### 8.1.3. Tested strategies

In order to evaluate the proposed approach, Monitoring Avoiding Dynamic Areas (*MADA*), and elucidate its advantages, we compare the proposed method with two other approaches: a riskier one, *Greedy*, and another more conservative one, *MADP*:

- *Greedy*. This approach entails the robot selecting the closest point of the frontier between the observed and non-observed regions. There is no an explicit representation of the dynamic areas to be avoided; the robot will navigate among the obstacles, without considering the possibility of loosing monitoring parts of the scenario whilst traverses these areas. It also employs FMM to compute obstacle-free paths and utilizes the path follower to follow them. The wavefront is propagated, surrounding the visible dynamic obstacles and computing the distance gradient from the robot until it reaches the closest non-observed cell, which is established as the next goal. A simple gradient descent from the goal is used to obtain the path. This strategy entails a certain degree of collision risk, mainly in dense areas.

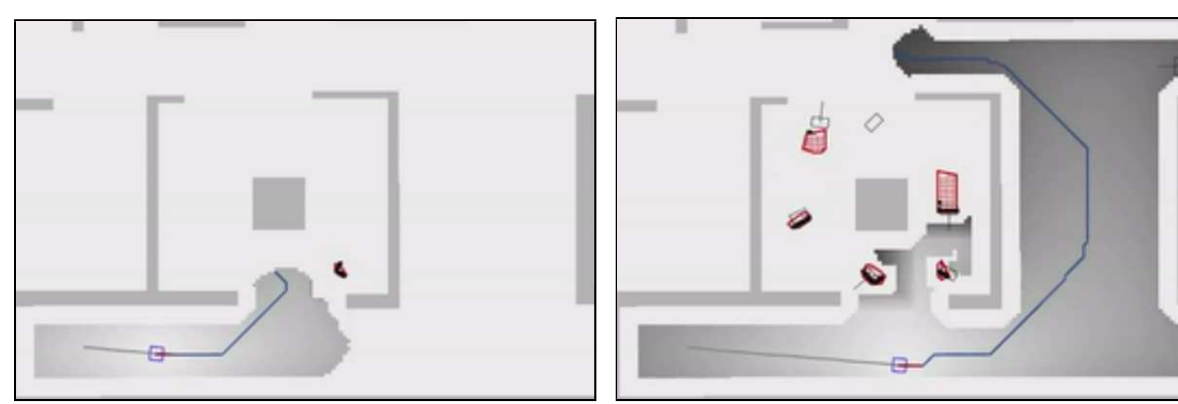

(a) Robot following the path (b) Update of the plan

Figure 17: Mission execution with *MADA*. In Fig.17(a), the robot follows the initial plan (blue path) to the selected goal. After observing moving obstacles, the robot updates the plan: building dynamic areas (red polygons), recomputing the goal in a new partition to monitor and a new path (blue) to reach it in Fig.17(b).

- *Monitoring Avoiding Dynamic Partitions (MADP).* Instead of building dynamic areas when the robot detects moving obstacles in a partition, the entire partition is considered dynamic, allocating a cost to it. The risk of collisions is reduced, but at the cost of entire partitions remaining unobserved. The robot could encircle these areas, resulting in a greater delay in mission time. This approach is a *conservative* variation of *MADA*.

With *MADA*, the robot exhibits a *balanced* behaviour between the *Greedy* technique and the more conservative *MADP*, as illustrated in Fig.17. According with the following evaluation based on the metrics, *MADA* can monitor most of the scenario, similar to the *Greedy* approach, reducing the risk of approaching dynamic obstacles and so of the collisions. Additionally, it should achieve this without a substantial increase in distance traveled or mission time, as observed with the conservative *MADP* approach, which causes longer detours.

In *MADP* and *MADA* approaches the robot will decide whether or not to traverse dynamic areas based on a occupancy threshold $O_{th}$ set by the user. So that, if a unique way to access to a goal is traversing a dynamic area whose obstacle density exceeds the threshold $O_{th}$, the goal will be discarded. All partitions corresponding to the branches of the tree traversing these dynamic areas are considered unreachable and discarded as well. Otherwise, if the density is less than $O_{th}$, the goal and the partitions are considered reachable. Therefore, the robot will take the risk of traversing the dynamic areas.

#### 8.1.4. Metrics

We considered the following metrics: (i) *traditional metrics*: the total path length traveled by the robot during the monitoring mission, the total time required by the robot for monitoring, the occurrences of collisions during the mission, and the computation time; (ii) *task context metric*, such as the ratio of the observed space relative to the free space for different density and velocity of moving obstacles in the scenario; (iii) *social behaviors metrics*, qualitatively evaluated by means of the distance from the robot to the obstacles, balancing the risk of collision, the space clearance, the path length, and the potential disturbance produced to the moving group. Collisions represent a ”hard” criterion, as a collision equates to a failure and ends the mission. In contrast, the distance to obstacles is a ”soft” criterion, reflecting how much the robot may disrupt humans moving in the environment.

### 8.2. Simulations

The proposed method was implemented in C++ on the Robotic Operating System (ROS). The simulations were conducted using the Stage simulator on a machine equipped with an Intel Core i7-13700 processor running at 2.1GHz with 16GB of RAM. Multiple simulations were launched, varying the previously described parameters to thoroughly evaluate the performance of the method: strategies (*Greedy*, *MADA*, *MADP*), occupancy threshold $O_{th} = [0, 10, 100]\%$, safety distance based on the robot radius $[R, 1.5R]$, two scenarios [Fig.16(a), Fig.16(b)], and the number of obstacles and their velocities, as described in Sect.8.1.1 and 8.1.2.

#### 8.2.1. Strategies

Figure 18 shows the overall results of all simulations for the three evaluated strategies in terms of all the metrics defined in Section 8.1.4. Regarding the context of the mission, the main metrics to observe are the % of monitored scenario and of collisions, and the distance to obstacles social metric. The other metrics are not so important in this context, and are a consequence of the decisions made to reach the more relevant metrics.The analysis yielded the following main conclusions, based on the mean, median and quartile values obtained.

- Regarding Figs. 18(a) and 18(b), we can conclude that *MADA* is able to monitor a similar percentage of the scenario that *Greedy*, but with fewer collisions. The *MADP* method results in a lower number of collisions, but the monitored space is significantly reduced by approximately 20% because *MADP* discards entire partitions. *MADA* is the balanced version, achieving more organized monitoring (thanks to the high-level planner) and avoids obstacle potentially-collision zones.

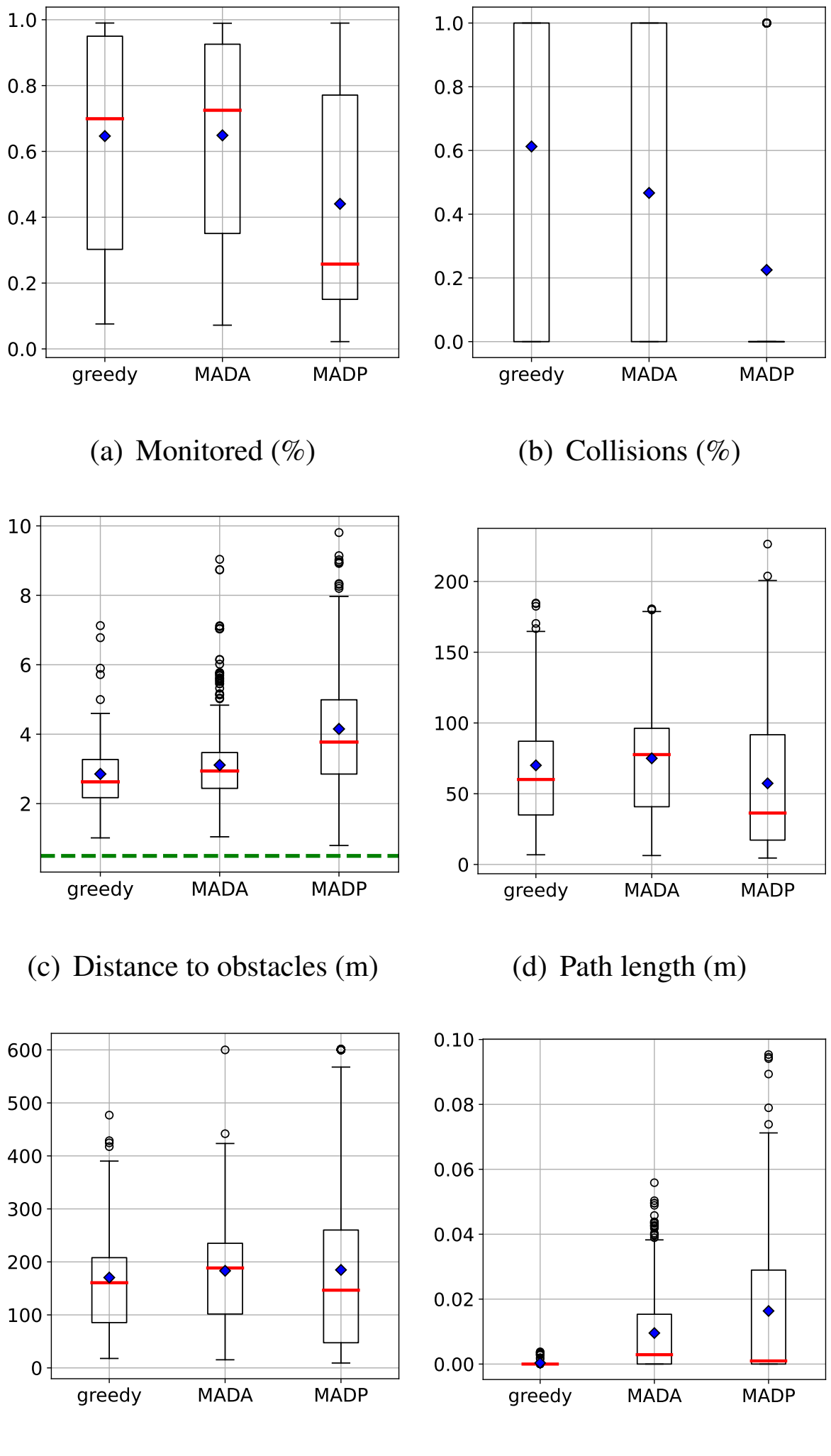

Figure 18: Total results based on the employed strategy.

- Taking into account the distance to obstacle metric, Fig.18(c), *MADA* lightly increases this distance with respect to *Greedy*, being bigger for *MADP*. In *Greedy* approach, the robot navigates close to the obstacles, increasing the rate of collisions, whilst *MADA* and *MADP* try to avoid the areas occupied by moving obstacles, traversing the areas with lower density, therefore reducing the collision risk. Obviously, the distance to obstacles increases with *MADP* because the robot is enforced to moving away from dense dynamic areas included in the avoided partitions.

- With *MADP* and *MADA*, the robot avoids partitions with moving obstacles, but is enforced to traverse partitions with lower obstacle density if no path through obstacle-free regions is available, resulting in *Greedy*-like behavior. Additionally, *MADA* extends and merges dynamic areas based on obstacle movement. Therefore, in the case of an obstacle moving towards a static or another dynamic obstacle, *MADA* may obtain paths in the direction of obstacle's movement, causing collisions. This problem has a possible solution, which is presented in Sect.8.3.1.

- Path length, Fig.18(d), and mission time, Fig.18(e), metrics are related. The robot travels more distance with *Greedy* and *MADA*, because with both techniques it monitors most of the scenario. In contrast, with *MADP*, the robot travels on average 25% less distance than with *MADA*, leaving regions to be seen, but spending a similar mission time. These classical metrics are not relevant in this mission context, in the case of *MADP* are a consequence of a lower monitored space than with the other methods.

- Finally, computation time, Fig.18(f), is lower in *Greedy*, which does not neither compute dynamic areas nor high and low level plans. However, even in the worst case, any variation of the proposed technique, *MADP* or *MADA*, computes the solution in less than 100ms, and can by applied in real time.

Evaluating the results by combining all the metrics allows us to understand the general trend of the proposed *MADA* method, including both its overall weaknesses and strengths. In the following subsections, we identify the best combination of metrics to be used in real-world experiments, in Section 8.5.

#### 8.2.2. *Occupancy threshold*

The threshold $O_{th}$, a parameter of the method, represents a dynamic area density value to decide if a such area can be traversable or not. Three values has been tested: 0%, in which case the area is not traversable; 10%, lower densities allow to traverse the area navigating the robot among the obstacles; 100%, the area is traversable, it is not avoided by the planner.

Based on the results of monitoring, Fig.19(a), and collisions, Fig.19(b), it is better to use *MADA* or *MADP* with $O_{th} = 10\%$, because both approaches achieve a good balance between monitored part of the scenario and with lower number of collisions than *Greedy*.

When the occupancy threshold is not zero, *MADA* shows no significant difference in the monitored area,

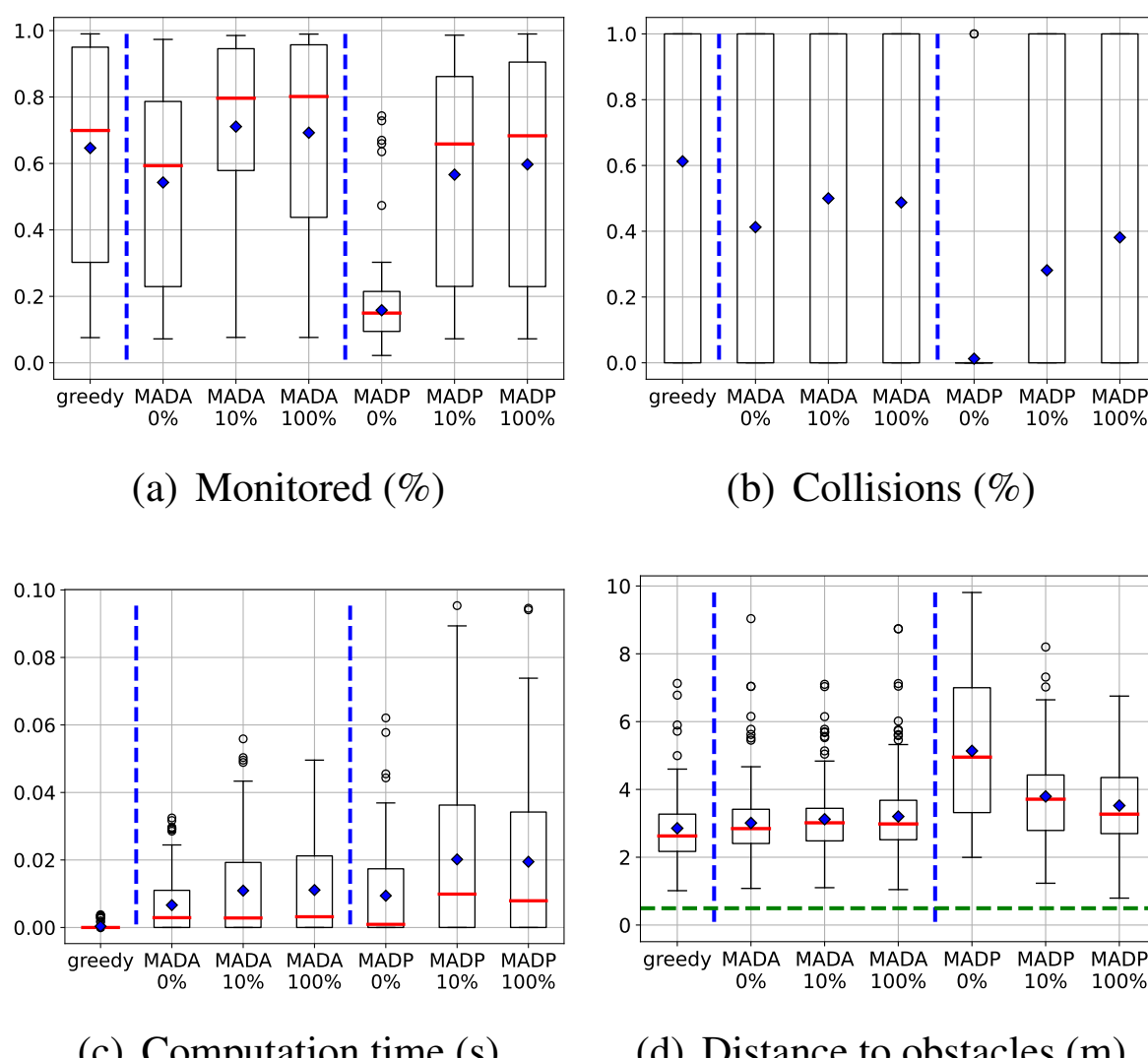

(a) Monitored (%) (b) Collisions (%)

(c) Computation time (s) (d) Distance to obstacles (m)

Figure 19: Results based on the occupancy threshold.

collisions, or distance to obstacles. This is because rapid expansion of dynamic areas due to obstacle movement quickly reduces density, making areas more traversable and prioritizing less dense ones.

When $O_{th} > 0\%$, it does not affect the computation time, Fig.19(c), because the costly part is the construction of the area itself more than the processing of the density.

### 8.2.3. *Safety distance*

A safety distance is added to the obstacles and the dynamic areas to enlarge them, increasing the safety distance from robot to obstacles. As can be seen in Fig.20, the monitored area is increased and collisions are reduced with the inflation. However, *MADA* keeps the best balance for both metrics, significantly increasing the monitored space (10%), but strongly reducing the rate of collisions (35%).

### 8.2.4. *Handling of dynamic areas*

In this section, we have evaluated four different ways the robot handles information when it sees again the previously constructed dynamic areas: (i) memorize everything, including both the points of the trajectories and the areas constructed from them; (ii) keep the points, but forget the areas if there is no obstacle inside when are seen by the robot (if there is an obstacle inside, the area is restored); (iii) forget everything, both points and areas, if there are no observed obstacles inside them within the visibility range; (iv) forget the areas in LoS (Line-of-Sight), but keep the areas that are not seen, having a *Greedy* behaviour in LoS.

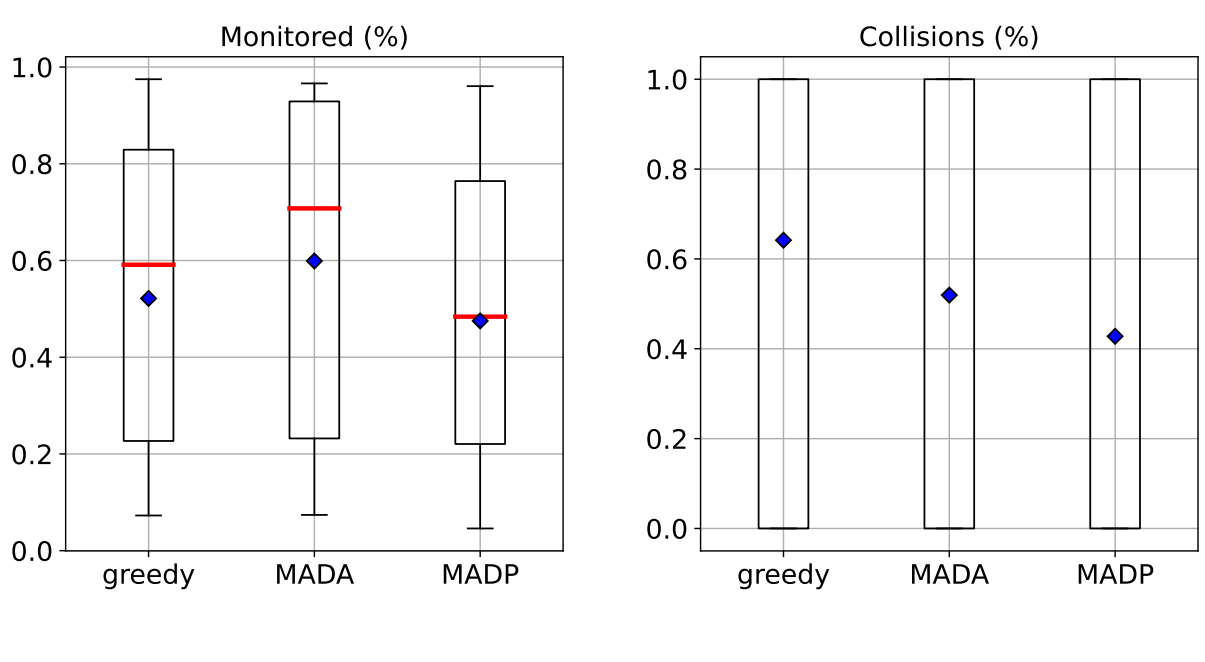

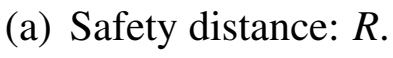

(a) Safety distance: $R$.

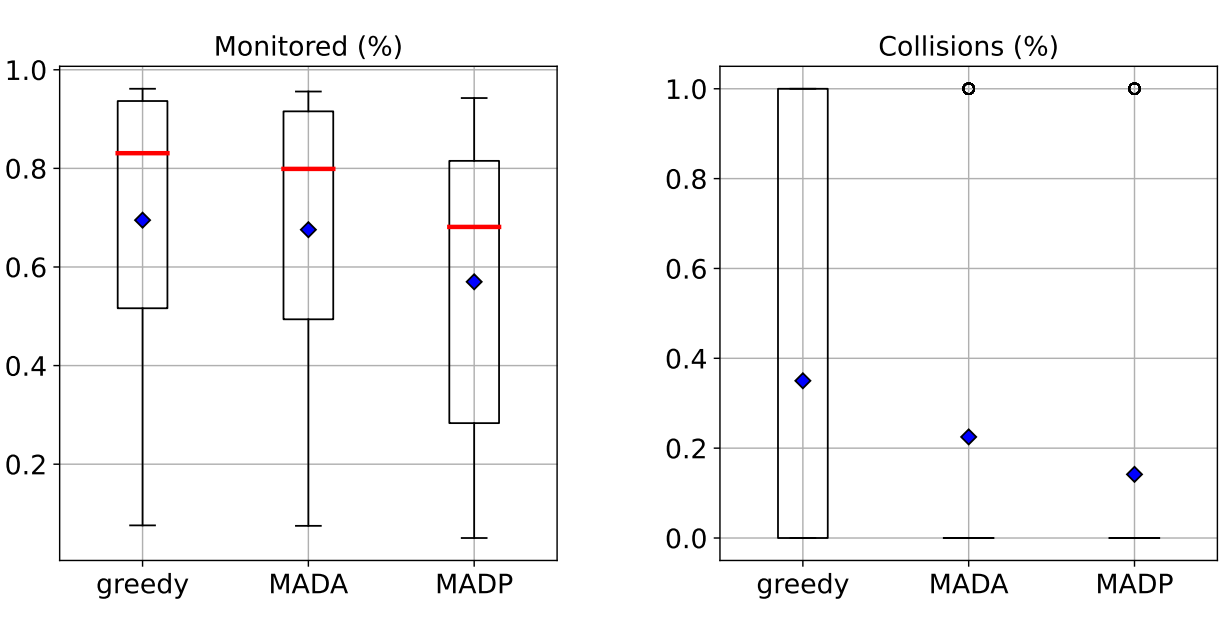

(b) Safety distance: $1.5R$.

Figure 20: Results for different safety distances.

This way we obtain different behaviours of the global planner. (i) and (ii) keep track of the detected dynamic areas, to take them into account for a further re-planning; the first one keeping the whole areas, the second one leaving more room for re-planning, but being aware that in these areas were occupied by moving obstacles. On the contrary, (iii) and (iv) forget the whole information of the dynamic areas detected when they are not seen, leaving more space for re-planning when needed; building and using the first one the dynamic areas for planning paths, and using a greedy technique for navigation among the obstacles. We aim to determine what is the best one in terms of the defined metrics.

The results are shown in Fig.21. Again, *MADA* and *MADP* outperform the *Greedy* approach in the ratio of collisions and monitored space. *MADA* improves the monitored area against *MADP*.

Regarding *MADA*, (ii) and (iv) strategies are the best; it is advantageous to keep storing trajectories and forgetting areas when the robot sees no obstacle in the area, or forgetting them when they are in LoS. This behaviors achieve the best ratio of monitoring and collisions, outperforming both *Greedy* and *MADP*. The robot effectively monitors the environment while avoiding situations where obstacles push it towards walls, thus re-

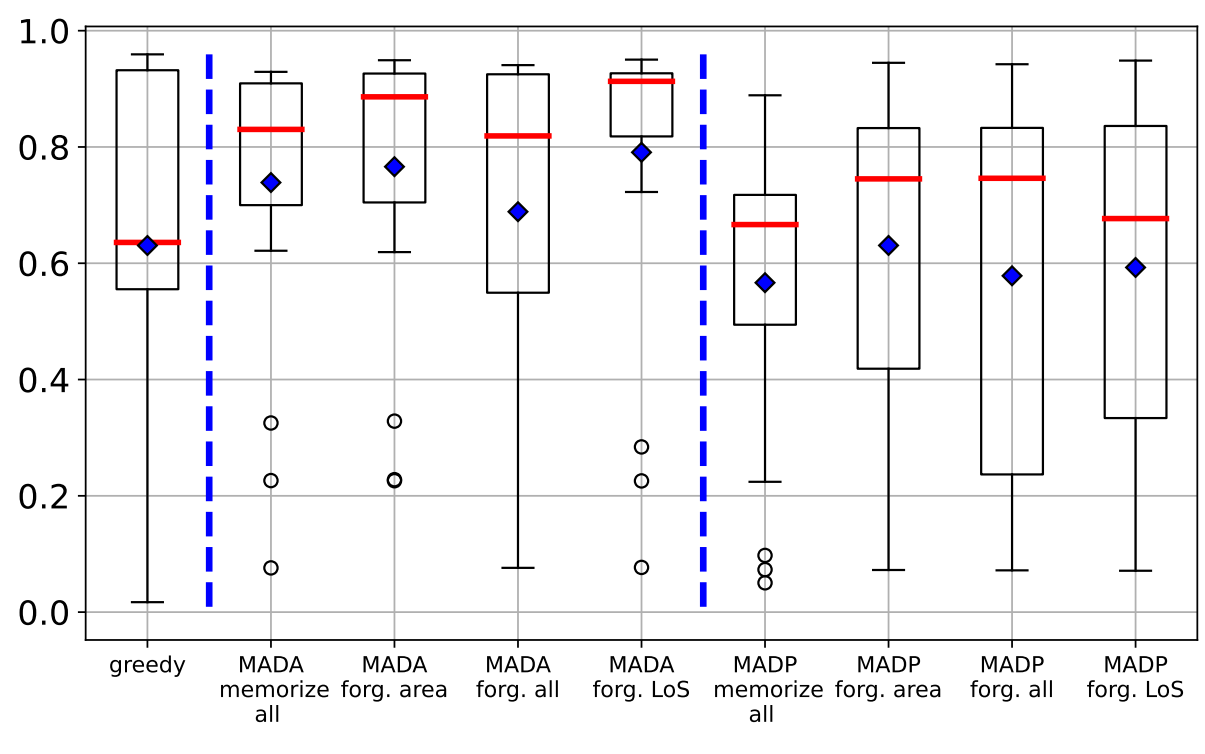

(a) Monitored (%)

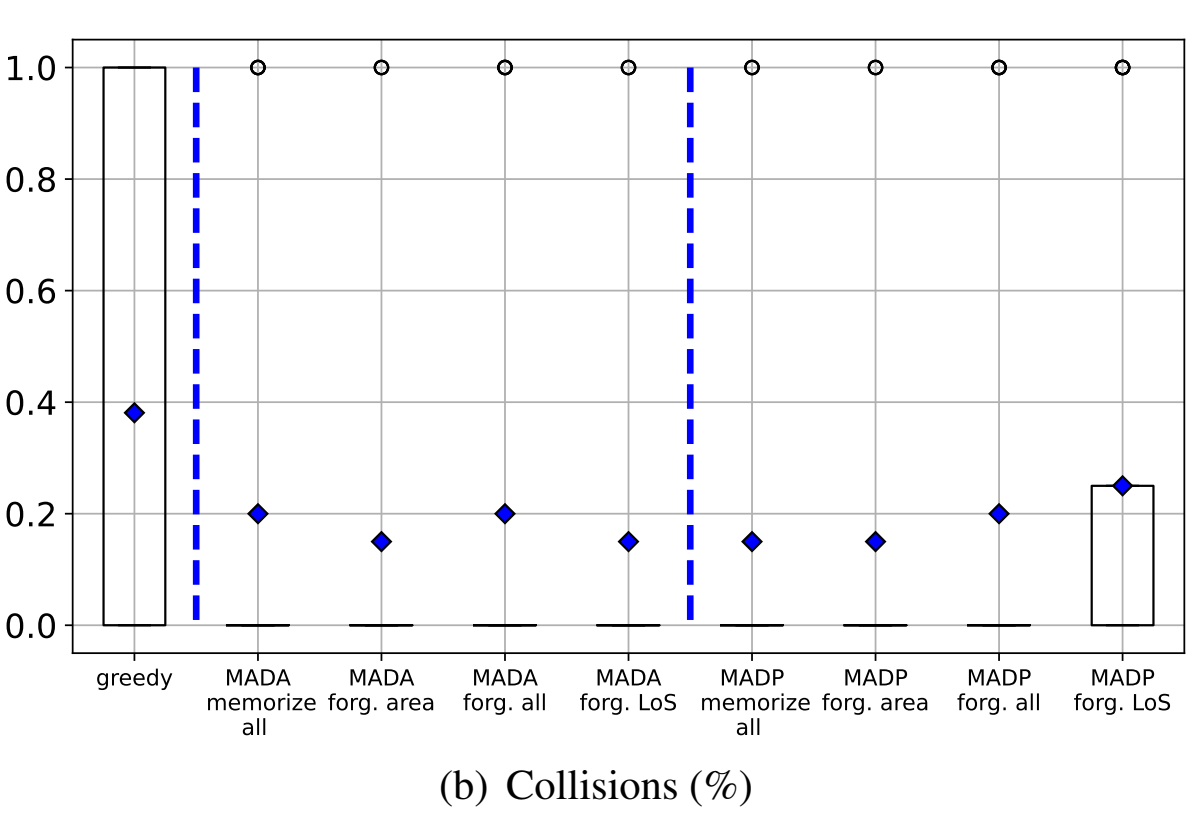

(b) Collisions (%)

Figure 21: Results based on the handling of the information of the dynamic areas.

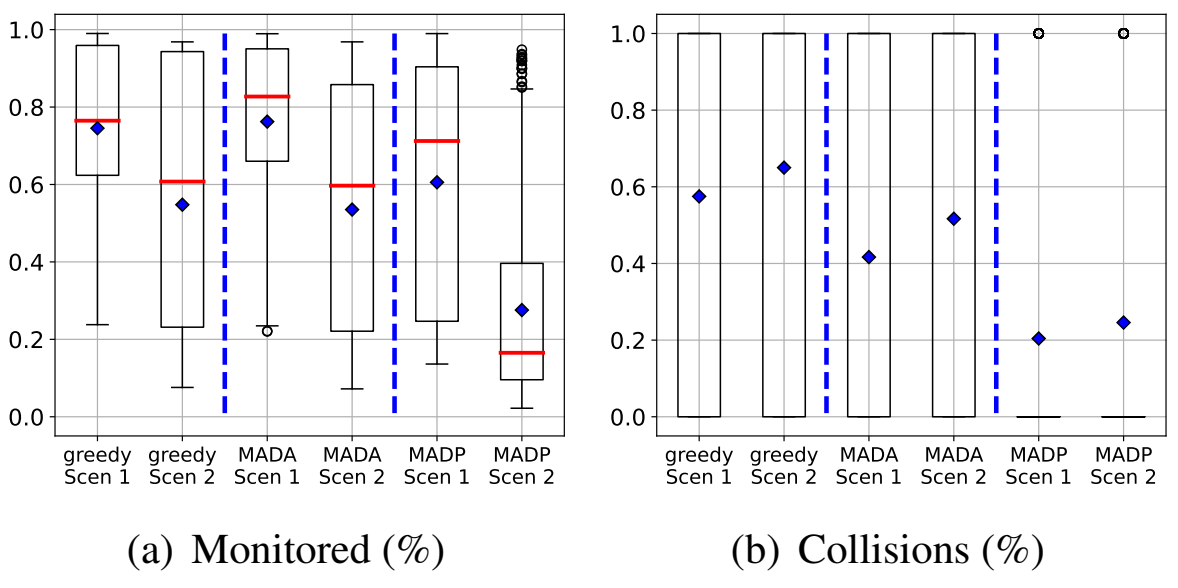

(a) Monitored (%) (b) Collisions (%)

Figure 22: Results based on the scenario.

ducing collisions.

### 8.2.5. *Scenarios*

The Fig.22 presents the results for the tested techniques in both scenarios, described in Sect.8.1.1. The rate of collisions is higher and the monitored scenario lower for all the methods in the second scenario. This is more challenging for monitoring missions due to its large stretch corridor and wide open spaces where dynamic obstacles can freely move.

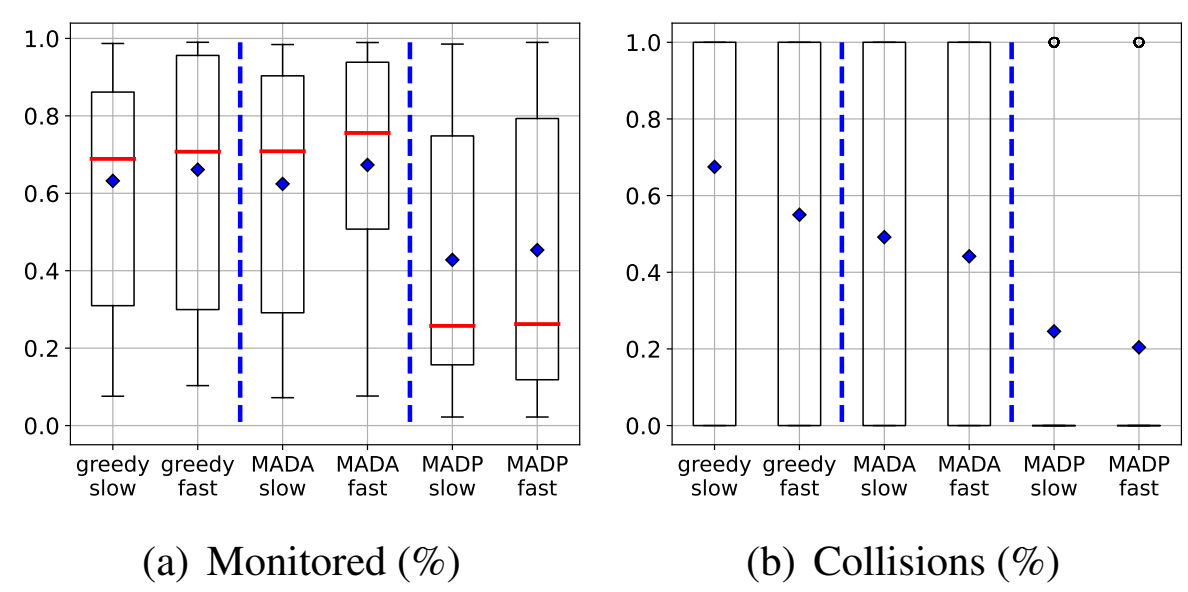

(a) Monitored (%) (b) Collisions (%)

Figure 23: Results based on the velocity of the obstacles.

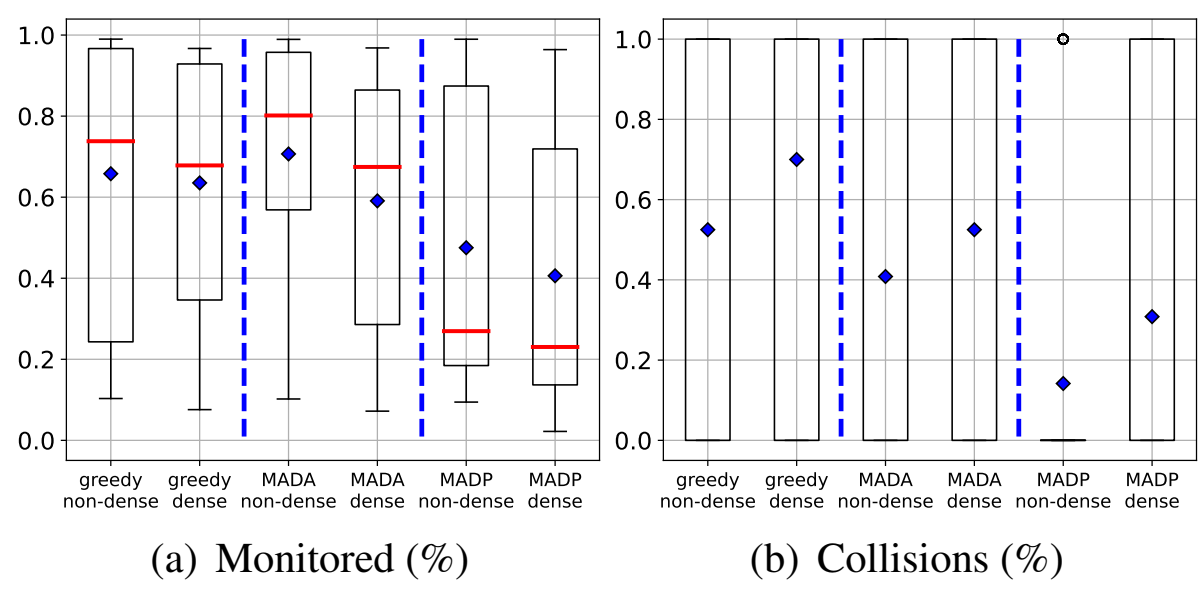

(a) Monitored (%) (b) Collisions (%)

Figure 24: Results based on the number of obstacles.

Computational time is higher in the second scenario, because of its size. But all strategies remain within the 100ms threshold, considered real-time. This is because the high-level planner selects the closest goals, and the low-level planner expands the FMM wavefront only to these goals.

### 8.2.6. *Velocity of obstacles*

Across all methods, it is observed that as obstacles move faster, more space is effectively monitored and fewer collisions occur, see Fig.23. This outcome is primarily influenced by the nature of the obstacle movement rather than the methods themselves. These movements, characterized by sudden changes from stationary to movement, pose challenges in prediction. These complexities are highlighted and better illustrated in the video.

### 8.2.7. *Number of obstacles*

The number of obstacles determines the density of environment where the robot is monitoring: dense and non-dense environments. We illustrate the results for both densities in Fig.24. As expected, the denser the environment, the less it can be monitored and the more collisions occur. *MADP* results in very few collisions, especially in low-density environments, but it typically

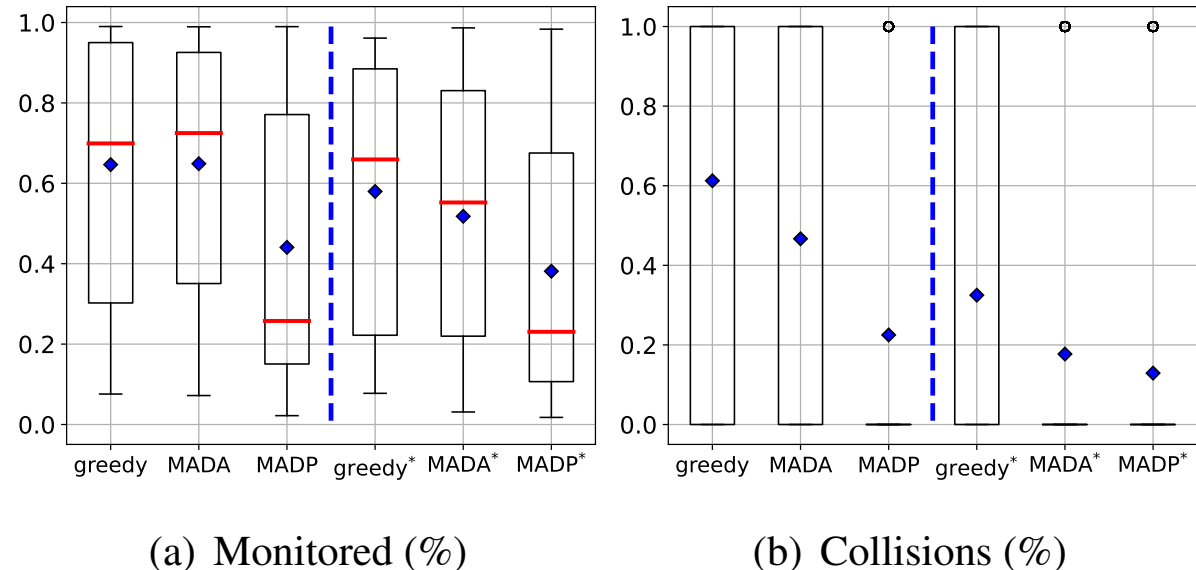

(a) Monitored (%)

(b) Collisions (%)

Figure 25: Results adding the projections of future positions of the obstacles. The strategies with the projections are denoted with $*$.

monitors less than 50% of the scenario, as it avoids and discards many regions. On the other hand, *MADA* has a higher collision rate, but it consistently manages to monitor nearly 60% of the scenario in non-dense environments.

### 8.3. Modularity for extensions

We aim to highlight the modularity of the proposed method, highlighting how easily new modules can be integrated or existing ones replaced at any stage of the diagram shown in Fig.3 of the Section 4.

A clear example of this modularity is seen in the techniques evaluated in the present work. The complete technique proposed is *MADA*, but in fact *Greedy* and *MADP* exemplify the modular nature of the method. *Greedy* does not use the high-level planner or the construction of dynamic areas. Meanwhile, *MADP*, is exactly identical to *MADA* except for using a different approach to construct dynamic areas.

#### 8.3.1. Projection of future obstacle movements

As previously described, one of the problems we encountered is when a dynamic obstacle creates a dynamic area, pushing the robot into a wall as it tries to navigate through free space. A simple solution implemented here is to predict future positions based on the velocity of the obstacles detected by the robot, adding greater safety. This module is inserted right before the low-level planner in Fig.3.

The results of the different strategies are illustrated in Fig.25. There is a significant improvement for all three techniques. All of them reduce the number of collisions by 50% compared to the techniques without projections. The improvement is even more significant for *MADA* versus *MADA*$^*$. It comes at the cost of exploring an average of 10% less of the scenario compared to the version without projections. It is arguably advantageous to add such predictions. More complex prediction methods could be implemented, but they are out of the scope of this work. The objective was to analyse how taking into account the future position of the obstacles using a simple technique, influence the metrics.

#### 8.3.2. Other planners

In this work, we have presented a global planning technique for dynamic environments. We assessed our method with and without using the high-level planner, greedy and *MADA* in the Fig.26, respectively. We do not evaluate reactive navigators alone for maneuvering among obstacles. Instead, we evaluate and compare our proposal with other well-established state-of-the-art global planners integrated with different reactive planners. We assessed our method, by replacing different parts of the low-level planner:

- Using the `move_base` node[2] (mb in the results). This node handles everything. The high-level planner sends the goal, and `move_base` proceeds to plan the path using Dijkstra and generates the appropriate speed commands to follow it using the Dynamic Window Approach (DWA)[40].
- Dijkstra+DWA(dd in the results). Similar to the `move_base` node, but we update the costmaps and plan with Dijkstra and DWA by ourselves, instead of `move_base`, generating the optimal path and the best speed command at each instant.
- FMM+DWA (fd in the results). Uses our path planner (FMM) to generate paths that avoid dynamic areas and employs DWA to follow the path.
- The *MADA* approach FMM+path follower (fp in the results). Uses FMM for path planning avoiding dynamic areas and our path follower for execution.

A qualitative advantage of our methods is the absence of parameters requiring adjustment for operation. The only user-set value is the occupancy threshold, which determines obstacle density for traversable dynamic areas. Except for this parameter, the planners self-adjust according to the environment, considering the robot's size and distance to obstacles.

Regarding quantitative values, shown in Fig.26:

- Using `move_base` is not effective, as it does not control the planning process, causing the robot to get stuck and fail to explore over 50% of the scenario, despite having the fewest collisions.

[2] `http://wiki.ros.org/move_base`

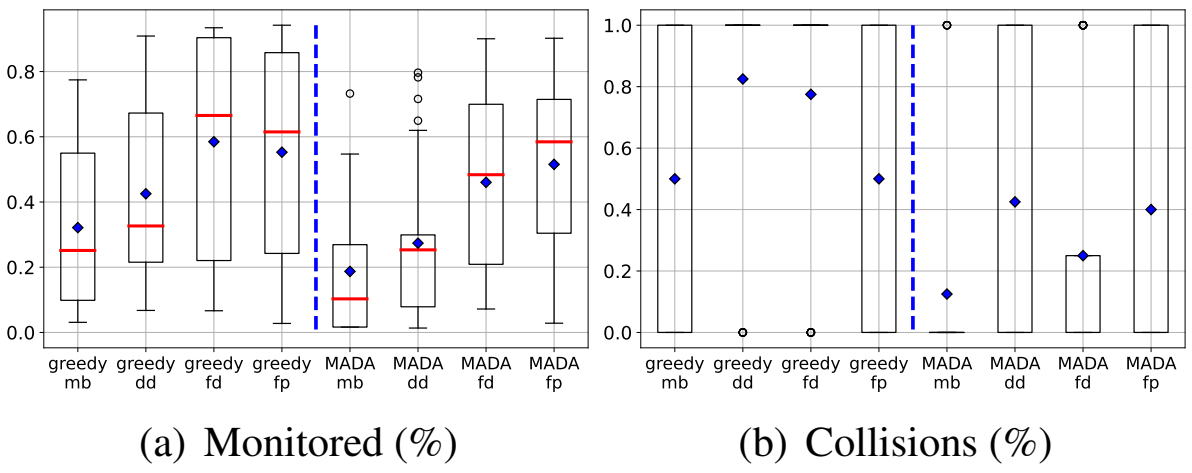

(a) Monitored (%) (b) Collisions (%)

Figure 26: Results based on the employed planners.

- The Dijkstra+DWA combination results in aggressive behavior with high speeds, leading to many collisions due to overlooked dynamic areas.
- Our path planner *MADA* improves the results. Both the method with the path-follower (fp) and its version with DWA (fd) achieve a better balance between scenario monitoring and collisions. FMM + path follower outperforms FMM + DWA with *Greedy*, monitoring 5% less but with 30% fewer collisions. Incorporating dynamic area construction and avoidance, FMM+DWA collides 15% less, and FMM+path follower monitors almost 10% more. This shows that once dynamic areas are identified, DWA reacts better to sudden movements, avoiding collisions.

This evaluation concludes that: (i) a global planner for dynamic environments is essential, regardless of the integrated local planner or reactive navigator; (ii) a good local planner integrated with the global planner for dynamic scenarios improves the whole system. A standalone local planner, even if tailored for dynamic environments, is insufficient for optimizing or adapting to specific missions, as correct subgoal selection and handling dynamic areas are crucial for mission success.

### 8.4. Comparison with other methods

In this section we have compared the proposed method *MADA* with other state-of-the-art approaches in the monitoring mission. The first, is a method of avoidance of crowded regions. The second, is a social path planner employed in monitoring missions altogether with the dynamic region avoidance, proposed in this work.

#### 8.4.1. Crowded regions avoidance

As discussed in Section 2, most methods there mentioned focus on navigation or path planning in human environments. In [21], the authors propose the Crowd-Based Dynamic Blockages (*CBDB*) layer, which shares similarities with our approach by avoiding crowded regions that hinder navigation due to obstacle density. Using the *CBDB* layer, the robot evaluates potential paths and selects between navigating through or around obstacle-filled areas. This method resembles our *MADP* approach, where large regions are considered occupied or not based on a threshold. Thus, we compare the *CBDB* approach with our *MADA* method, which dynamically adjusts dynamic areas based on obstacle movement. The high-level planner generates a goal from which the robot makes the next observation. Then, the *CBDB* layer is applied to define and evaluate the regions with possible blockages. For a fair comparison, we use FMM to compute the path, instead of A* used in [21], in order to maintain the robot far away from static obstacles and maximize the field of view, as explained in previous sections. And both strategies use DWA for navigation. The comparison focuses on how dynamic areas are handled. With *CBDB*, large regions around moving people are blocked, while with *MADA*, the areas are adjusted to the actual space occupied by obstacles, allowing more space for navigation. Simulations were conducted in the scenarios shown in Fig.16, with results illustrated in Fig.27.

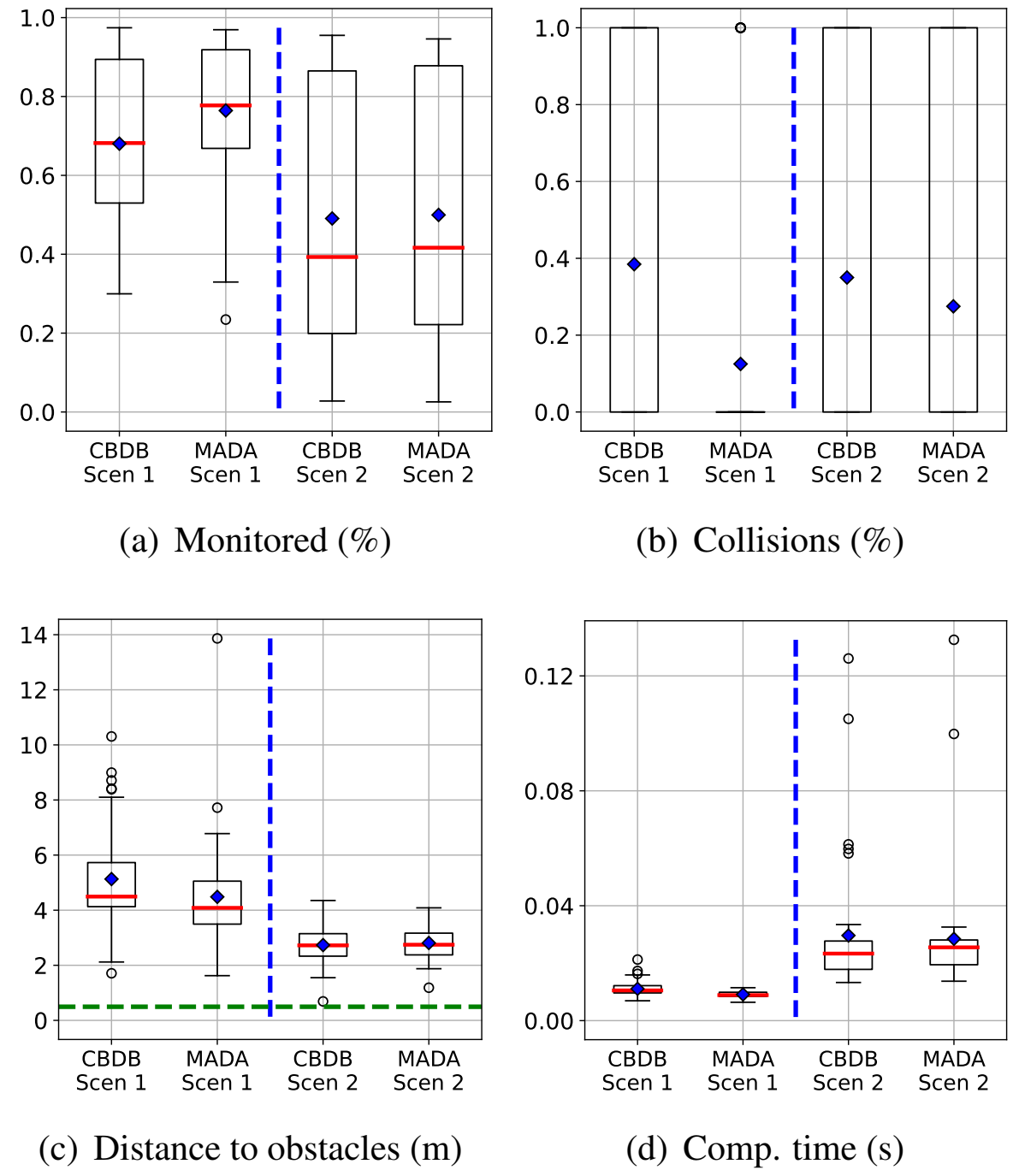

(a) Monitored (%) (b) Collisions (%)

(c) Distance to obstacles (m) (d) Comp. time (s)

Figure 27: Comparison with a method for crowded region avoidance.

As shown in Fig.27(a), the robot monitors a larger portion of the environment using *MADA* compared to *CBDB*. This is because, despite moving closer to dynamic obstacles (Fig.27(c)), *MADA* results in fewer collisions, in Fig.27(b). This suggests that considering large regions, especially in environments with dispersed obstacles, is less effective when all paths cross obstacle-dense areas. Instead, *MADA* creates regions that adapt to the observed movement of dynamic obstacles, allowing the robot to navigate through obstacle-free gaps and avoid collisions. In Fig.27(b), the difference in collisions is smaller in the second scenario due to its corridor-like structure, with fewer path options. This structure also affects computation time, in Fig.27(d), as *CBDB* evaluates fewer paths, leading to a faster computation than *MADA*. However, in the first scenario, with more bifurcations, *CBDB* takes slightly longer than *MADA* to compute the best path.

In conclusion, *CBDB* offers computational time savings in simpler environments with few bifurcations. However, in more complex environments with many corridors and rooms, *MADA* performs similarly in terms of time, and its dynamic region adaptation allows for collision-free paths by leveraging the movement of obstacles in more spacious areas.

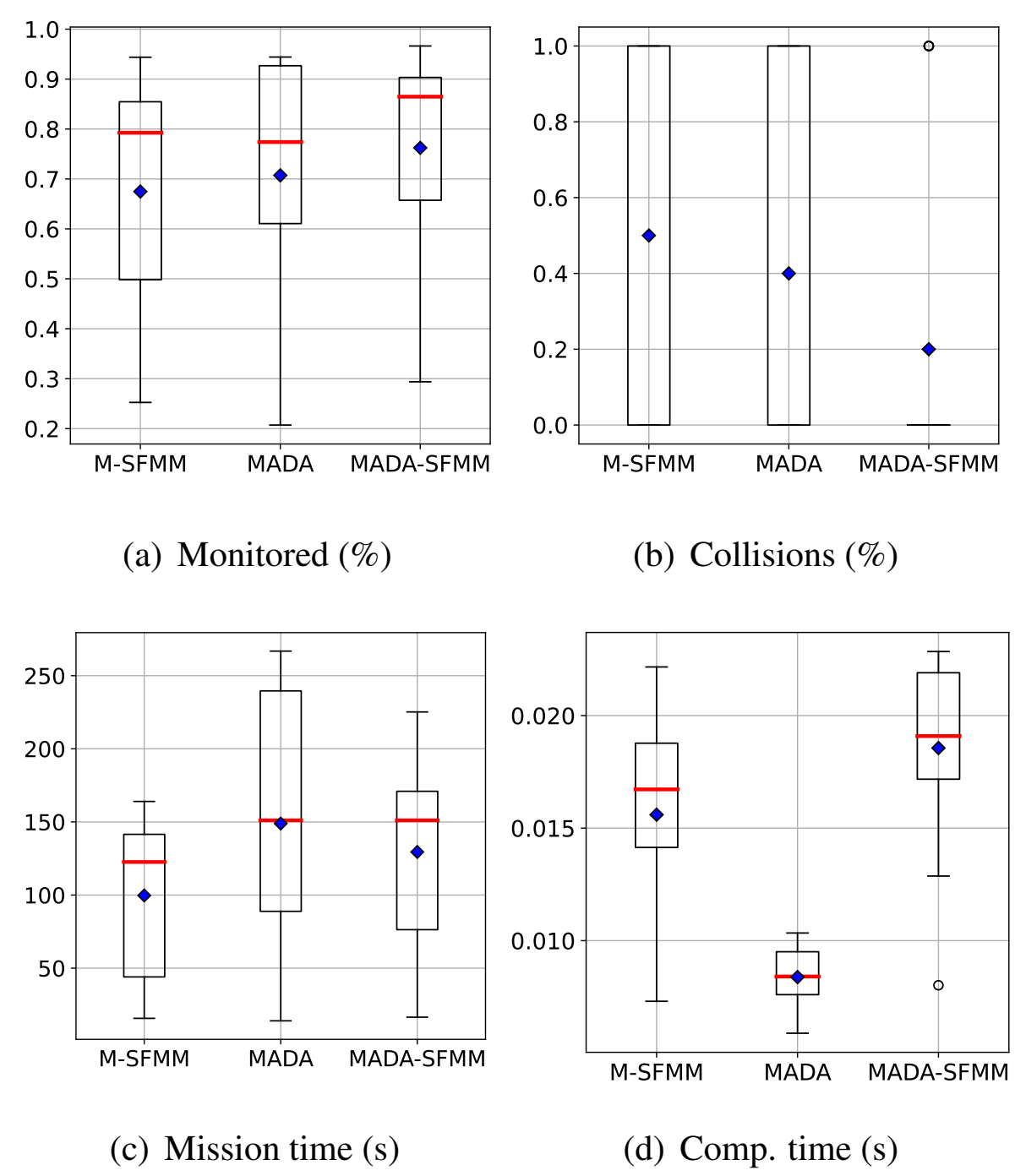

(a) Monitored (%)

(b) Collisions (%)

(c) Mission time (s)

(d) Comp. time (s)

Figure 28: Comparison with a social planning method.

### *8.4.2. Social path planning*

In this section, we demonstrate the benefits of integrating dynamic area information with social path planners. Specifically, we incorporate the social path planner from [16] (referred to as SFMM) into our global planner for monitoring missions involving dynamic obstacles. SFMM employs the Fast Marching Method (FMM) for social path planning in environments with humans. It incorporates a predefined *social space*, based on [41], using a Gaussian mixture model in the velocity map to generate a gradient that repels the robot's path away from people. SFMM is applied in various scenarios, such as navigating toward specific goals while avoiding humans, or interacting with individuals or groups. Our focus is on the former: a robot navigating through shared spaces with individual people.

We chose this planner for comparison due to its similarities with our approach. First, it is based on FMM, making integration into our method straightforward. Second, SFMM ensures safe trajectories by maintaining a safe distance from obstacles. However, there are two key differences from our proposed approach. SFMM trims the velocity map generated by static obstacles to maintain a certain degree of safety, without diverting the path too far from the optimal route to the goal. In contrast, our approach maximizes safety by keeping the paths as far from obstacles as possible, enhancing the robot's field of view for better monitoring. Additionally, we incorporate dynamic areas, which represent regions where obstacles are in motion.

Thus, we incorporated the SFMM planner into our approach to assess whether including dynamic areas offers advantages. Simulations were run in a high-obstacle-density scenario of Fig.16(a). For a fair comparison, all planners use DWA to navigate. The simulation results are presented in Fig.28. As before, the high-level planner, described in Section7.1, is responsible for generating monitoring goals, and the path planner computes the path to reach these goals. We compared the SFMM with and without dynamic areas, referred to as M-SFMM and MADA-SFMM, respectively, in the results. As shown in Fig.28(a), with M-SFMM, the robot monitored between 50% and 85% of the scenario in most trials. Using *MADA*, this ratio increased to 60-93%. When both methods were combined MADA-SFMM, that is the management of dynamic areas and the Gaussian-based model for the moving people, the ratio improved further, reaching 65-90%. This improvement improvement was directly related to the collision rates, Fig.28(b). With M-SFMM alone, despite avoiding *social spaces* of the obstacles, the robot collided in 50% of the trials, due to navigating through crowded regions. With MADA, collisions were reduced to 40%, as the planner guided the robot through less crowded areas.

However, in densely populated environments like ours, it is nearly impossible to find completely obstacle-free zones, which explains the remaining collisions. The best performance was achieved by combining both techniques, MADA-SFMM, reducing collisions to just 20%. This improvement was due to the dynamic area information being incorporated into the velocity map to compute the gradient, as described in eq.6 of Section 7.2.2, while SFMM kept the path away from obstacles within the dynamic areas, increasing safety by adding the *social space* buffer around the obstacles.

In terms of mission time, in Fig.28(c), we observe that while M-SFMM is more risky, navigating through denser areas allows it to complete the mission more quickly, ranging 40-140s in most trials. On the other hand, *MADA*, takes longer — up to 85 to 235 seconds — due to taking greater detours to enhance safety. However, combining both methods with MADA-SFMM reduces the mission time to 70-170 seconds. This occurs because the planner includes the *social space* around obstacles, balancing safety with efficiency. MADA-SFMM offers the best of both worlds: it is safer, monitors more effectively, and completes the mission faster, with the trade-off being slightly increased computation time, as depicted in Fig.28(d).

In conclusion, planning global paths by explicitly managing dynamic area avoidance with *MADA* proves to be more practical than focusing on individual obstacle avoidance in such obstacle-dense environments. Incorporating dynamic area information, as proposed in our work, alongside other social planners significantly improves the path-planning process in terms of safety when the robot is forced to navigate regions shared with dynamic obstacles. Additionally, this highlights the modularity of our approach, which can easily integrate other path planners.

### *8.5. Real-world experimentation*

The proposed approach was tested in the hall of the Ada Byron building at the University of Zaragoza. An example of the execution of the mission is available in the video[3]. The experiments were conducted using a Turtlebot equipped with a Hokuyo lidar (20m range, 180º field of view) and a camera for scenario monitoring, with a imposed visibility range set to $v_r$ = 5m. That is, the robot registers from the LiDAR dynamic obstacles up to 20m away, but needs to approach within 5 meters to correctly monitor the scene. The robot is localized with AMCL, and only observed dynamic obstacles are depicted on the known static map, illustrated with black dots in the video. Initially placed in the middle of the hall, the robot monitored the entire environment. The entrance to the building is on the left side of the scenario, resulting in higher dynamism in that region. As shown in the videos, the robot encounters the most difficulty navigating there. Based on simulated results, we set the best combination of parameters: $O_{th}$ = 10%, safety distance $1.5R$, and memorizing obstacle trajectories but forgetting them when no obstacles are inside, the (ii) mode explained in Section 8.2.4. Four real-world experiments were conducted, resembling simulations: dense and non-dense environments, and with slow and fast human movements.

As seen in the video, people usually avoid collisions with the robot. When they move much faster than the robot, as in the first experiment, it is impossible for the robot to avoid disturbing them completely. Fast foot movements also complicate building dynamic areas, which is a problem we will address in the near future. However, as illustrated in Fig.29, the robot successfully completed the mission by navigating around areas with a high presence of dynamic obstacles whenever possible.

With slower human movements, the robot can better build dynamic areas, as seen in the video. This allows it to navigate around these areas more effectively and avoid disturbing people. The robot also chooses different directions, regions to monitor and plans different paths based on the observed dynamism. This is clearly seen in the third and fourth experiments, Non-dense/Slow and Dense/Slow respectively, where the planner selects opposite monitoring directions.

The code used for this research will be made available upon publication acceptance. This will enable researchers to reproduce the results and facilitate further development in this field.

## 9. Conclusions

In this paper, we have presented a method for global planning and navigation in the presence of moving obstacles, tailored for monitoring mission contexts. The proposed approach *MADA* involves the robot constructing dynamic areas where obstacles are moving, avoiding areas considered dense based on their occupation. The method was evaluated by comparing it with two other approaches: a *Greedy* strategy and a more conservative version of the proposed method, *MADP*, which avoids larger spaces with moving obstacles rather than building dynamic areas around them. The results show that the proposed technique achieves a good balance between the monitored portion of the scenario, the time to complete the mission, whilst avoiding regions more densely occupied by moving obstacles to reduce the risk of collision. *MADA*, by constructing dynamic areas that are adjusted to the observed obstacle movement, proves to be more efficient than other techniques that either build larger regions based on the environment or directly adjust to each individual obstacle using a social model. We have shown the modularity of the proposed method, which is advantageous for extensions of the method. The evaluation of the results reveals that a global mission planner which integrates dynamic global trajectory planner and local planners improves the performance in terms of safety and social criteria associated to a specific mission.

[3]`https://bitly.cx/1Zuv`

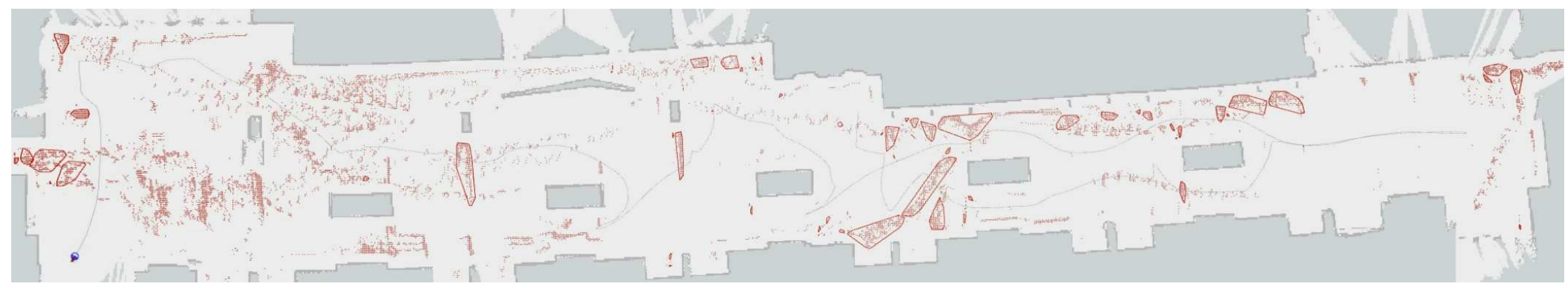

Figure 29: The robot accomplished the monitoring mission, traveling the path depicted by the gray line while avoiding regions with high occupancy of dynamic obstacles. The trajectories of people, registered by the robot (during the whole mission), are shown as red points, and the dynamic areas (at the end of the mission) are illustrated with red polygons.

The method can potentially be extended or adapted for use in other common robotic problems or applications different to monitoring, such as classical coverage problems or object-seeking tasks in scenarios with dynamic obstacles, only modifying the *High-level planner* for a context-aware execution. A potential future research direction is the extension of our method to multi-robot teams, introducing challenges related to partitioning and goal allocation among robots.

### Acknowledgments

This work was partially supported by the Spanish projects PID2022-139615OB-I00/MCIN/AEI/10.13039/501100011033/FEDER-UE and Aragon Government FSE-T45_23R.

### Declaration of generative AI and AI-assisted technologies in the writing process

During the preparation of this work the authors used Gemini in order to improve text readability. After using this tool, the authors reviewed and edited the content as needed and take full responsibility for the content of the published article.